\documentclass[english,11pt]{article}
\usepackage{geometry}
\usepackage{booktabs} % For formal tables
\usepackage{colortbl} % colored hline
\usepackage{amsmath}
\usepackage{amssymb}
\usepackage{xspace}
\usepackage{xcolor}
\usepackage{dsfont}
\usepackage{rotating}
\usepackage{float} % for [H]

\usepackage[algo2e,ruled,vlined,linesnumbered]{algorithm2e}
\usepackage{url}
\usepackage{amsthm}
\usepackage{amsmath}

\renewcommand{\epsilon}{\varepsilon}

\newcommand{\onemax}{\textsc{OneMax}}
\newcommand{\OM}{\textsc{OM}}

\newcommand{\ParadisEO}{{\fontfamily{lmss}\selectfont Paradis\textmd{eo}}}
\newcommand{\IOH}[1]   {{\fontfamily{lmss}\selectfont IOH{#1}}}
\newcommand{\irace}    {{\fontfamily{lmss}\selectfont irace}}

\DeclareMathOperator{\flip}{flip}

\DeclareMathOperator{\Evals}{Evals}

\DeclareMathOperator{\AUC}{AUC}

\usepackage{appendix}

\usepackage{color}
\definecolor{darkgreen}{rgb}{0.09, 0.45, 0.27}
\definecolor{darkred}{rgb}{0.55, 0.0, 0.0}
\definecolor{darkblue}{rgb}{0.0, 0.28, 0.67}

\usepackage{hyperref}
\hypersetup{
    colorlinks = true,
  linkcolor = darkred,
anchorcolor = black,
  citecolor = darkgreen,
  filecolor = cyan,
  menucolor = red,
   runcolor = cyan,
   urlcolor = darkblue,
}

\title{Towards Large Scale Automated Algorithm Design\\ by Integrating Modular Benchmarking Frameworks}

\author{Amine Aziz-Alaoui, ISAE-SUPAERO, Universit\'e de Toulouse, France\thanks{This work was partially done during the M2 master internship of Amine Aziz-Alaoui at \'Ecole Polytechnique, Institut Polytechnique de Paris, CNRS, LIX, Palaiseau, France. He is now a PhD student at Institut de Recherche Technologique Saint Exup\'ery, Toulouse, France.}\\
    Carola Doerr, Sorbonne Universit\'e, CNRS, LIP6, Paris, France\\
    Johann Dreo, Thales Research \& Technology, Palaiseau, France\thanks{Corresponding author, \href{mailto: Johann@Dreo.fr}{Johann@Dreo.fr}.}
}

\begin{document}

\maketitle

\begin{abstract}
We present a first proof-of-concept use-case that demonstrates the efficiency of interfacing the algorithm framework ParadisEO with the automated algorithm configuration tool irace and the experimental platform IOHprofiler. By combing these three tools, we obtain a powerful benchmarking environment that allows us to systematically analyze large classes of algorithms on complex benchmark problems. Key advantages of our pipeline are fast evaluation times, the possibility to generate rich data sets to support the analysis of the algorithms, and a standardized interface that can be used to benchmark very broad classes of sampling-based optimization heuristics.

In addition to enabling systematic algorithm configuration studies, our approach paves a way for assessing the contribution of new ideas in interplay with already existing operators---a promising avenue for our research domain, which at present may have a too strong focus on comparing entire algorithm instances. 
\end{abstract}

% \maketitle

\sloppy

%\sloppy

\section{Introduction} 
When confronted with an optimization problem in practice, one of the major challenges that we face is the selection (and the configuration) of an algorithm that corresponds well to the given problem structure, optimization objective(s), and the available resources (compute, possibility to parallelize computations, accessibility of the problem, etc.). A vast amount of different optimization techniques exist, which renders this \emph{algorithm selection problem} non-trivial. 

In practice, algorithm selection is often biased by personal preferences and experiences, as well as by practical aspects such as the availability of ready-to-use implementations. 
Supporting practitioners in making more systematic choices is one of the key objectives of our research domain. A key tool for deriving such recommendations is \emph{algorithm benchmarking}, \emph{i.e.,} the analysis of empirical performance data and search trajectories of one or several algorithms on one or several optimization problems~\cite{TBB20benchmarking,hansen2016coco}. Several important benchmarking tools and software frameworks have been developed by our community to ensure sound and meaningful data extraction. These platforms address different stages of the algorithm selection process. They cover, for example, instance selection and generation~\cite{instance15,Wmodel,surrogateBenchs}, feature extraction~\cite{flacco}, algorithm configuration~\cite{SMAC,SPOT,irace,hyperband}, experimentation~\cite{hansen2016coco,nevergrad,IOHprofiler,HeuristicLab}, data analysis~\cite{Borja19,EftimovPK20DSCtool,eaf}, and performance extrapolation~\cite{KerschkeHNT19}. However, most of these tool are  developed in isolation, paying little attention to building compatible interfaces to other benchmarking modules. This significantly hinders their wider adoption. 

With this work we demonstrate the benefits of a fully modular benchmarking pipeline design, which keeps the different steps of the benchmarking study in mind. We see our work as a proof of concept for better compatibility between benchmarking software. On the practical side, our pipeline paves a way for assessing the benefits of new algorithmic ideas in the context of and in interplay with other operators and ideas that our community has to offer. 

\subsection{Our Contribution}
% \textbf{Our contribution:} 
Concretely, we propose in this work a benchmarking pipeline that integrates the modular algorithm framework \ParadisEO{}~\cite{ParadisEO,ParadisEOjournal} with the algorithm configuration tool \irace{}~\cite{irace}, the experimental platform \IOH{experimenter}~\cite{IOHprofiler}, and the data analysis and visualization module \IOH{analyzer}~\cite{IOHanalyzer}. 
We test our pipeline on tuning a family of genetic algorithms, inspired by~\cite{YeWDB20}, on the so-called W-model problem instances suggested in~\cite{WmodelInstancesASoC}.

\emph{Quality of the results:}  We show that \irace{} is capable of finding algorithm instances which outperform all baseline algorithms selected by hand, and this for each of the 19 problem instances that we consider. The relative advantage of the best out of 15 \irace{} suggestions over the best baseline algorithm, measured in terms of volume under the discretized Empirical Attainment Function (see Sec.~\ref{sec:ECDFlogger}), varies between 1\% and 30\%, with a median gain of 13\%.

\emph{Scalability:} Targeting per-instance algorithm design on synthetic benchmark, our algorithmic framework is capable of generating large set of solvers, up to several millions of unique configurations. We show that it is possible to tackle such spaces thanks to fast computations. For instance, we give \irace{} a budget of 100\,000 target runs for each of 19 problems, and it completes the full task in approximately 3 hours on a laptop.
In our experience, our C++ pipeline is at least 10 times faster than heavily optimized counterparts in Python, not mentioning that most of the available modular frameworks are not always heavily optimized. 

\emph{Take-away for instance selection:} As a side result, we observe that similar algorithm instances can be suggested by \irace{} for some problems, suggesting that the diversity in performance profiles sought in~\cite{WmodelInstancesASoC} may be weaker than intended. Our work suggests that an approach like ours may result in a more reliable instance selection, since it will be less biased by a small set of baseline algorithms, but rather be built on a large and diverse set of possible algorithm instances.

\emph{Extendability:} Our pipeline is ready to perform large benchmark studies, covering large classes of continuous and discrete optimization algorithms. For example, local searches, particle swarm optimization, estimation of distribution algorithms and using numerical or bitstring encodings. Similarly, the pipeline gives direct access to all problems collected in \IOH{profiler}, which comprises in particular the BBOB functions from the COCO framework~\cite{hansen2016coco}, the Nevergrad problem suites~\cite{nevergrad}, the W-model instances~\cite{Wmodel}, the PBO suite~\cite{DoerrYHWSB20}, etc.
Additionally, any benchmark or solver which would be plugged into \IOH{profiler} would be easily used to further extend this study.

\subsection{Comparison to Previous Works} 
Our work is a top-down approach for automatic algorithm design~\cite{GrammarSLS}, which uses a parametrized algorithmic framework to instantiate many {\em algorithm instances}.
Following~\cite{lopez2017automatic}, we observe that this differs from bottom-up ``grammar-based'' approaches~\cite{DBLP:conf/evoW/SouzaR18, DBLP:conf/cec/SouzaR18,DBLP:journals/eor/PagnozziS19} or Grammatical Evolution~\cite{ryan1998grammatical,lourenco_evolving_2012},
which allow for easily designed algorithm space, but complicates algorithm instantiation and optimization.
In our case, the width of the design space is already large and we target fast algorithm instantiation. We thus favor the top-down approach.
In this first study, we only consider categorical parameters, for the sake of implementation simplicity.

A similar approach to ours was suggested in~\cite{AutoMACO,BezLopStu2019ec,AutoMOEA} for multi-objective optimization. Those studies also use \irace{}, but the authors implemented their own modular algorithm frameworks that are restricted to multi-objective optimization. 
Our work significantly scales up this kind of study, by leveraging larger algorithm design spaces, larger sets of benchmarks, with more problems and allowing a more detailed analysis of the results.

Few other studies consider a bi-objective measurement of performance for automated algorithm design.
Most notably, \cite{DBLP:journals/eor/Lopez-IbanezS14} introduced the use of the hypervolume for given quality and time budgets. In our case, we use the volume under the curve of the empirical cumulative histogram of quality and time attainments~\cite{DBLP:conf/emo/FonsecaFH01}. This should behave as the sum of hypervolumes defined for a set of thresholds covering the whole domain, instead of a single one.

\subsection{Structure of the Paper}
Sec.~\ref{sec:pipeline} briefly introduces the individual modules of our algorithm design pipeline and how they interplay with each other. The use-case on which we apply this pipeline, as well as the experimental setup are summarized in Sec.~\ref{sec:usecase}. The results of our empirical analysis are described in Sec.~\ref{sec:results}. We conclude our paper in Sec.~\ref{sec:conclusions} with a discussion on promising avenues for future work. 

% \textbf{Availability of Code and Data:} 
\subsection{Availability of Code and Data}
The code and data used for this study have been archived at~\cite{aziz_alaoui_amine_2021_4729568}. %,furong_ye_2021_4727890,johann_dreo_2021_4727896}%
Up-to-date versions of the code are available at \url{https://github.com/jdreo/paradiseo} and \url{https://github.com/IOHprofiler/IOHexperimenter}.

\begin{figure}[t!]
\centering %[trim=left bottom right top, clip]
 \includegraphics[width=0.78\textwidth]{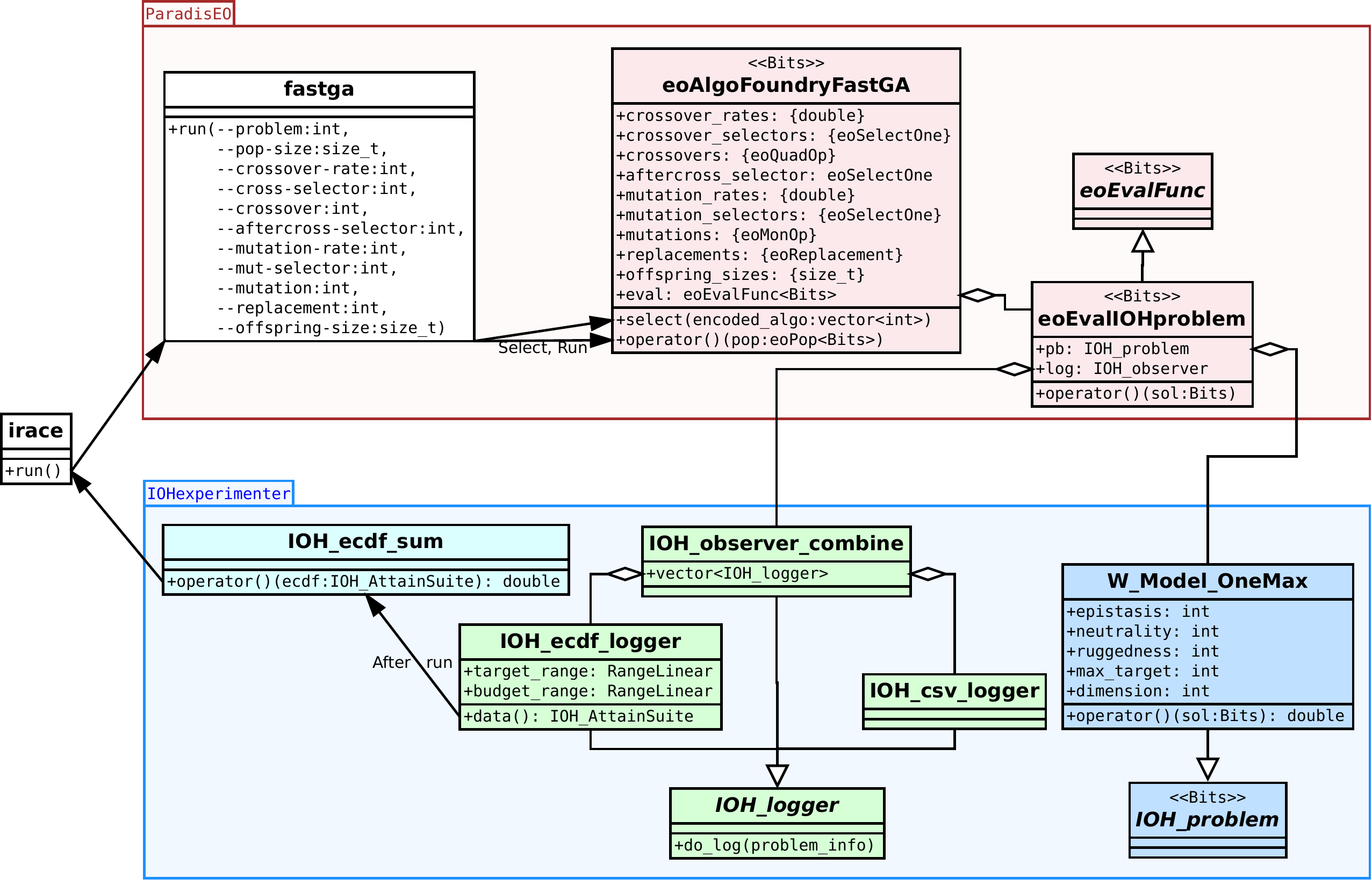}
\caption{Summary diagram of the FastGA evaluation pipeline involving the \ParadisEO{} (upper part, red colors) and \IOH{experimenter} (lower part, blue colors) frameworks along with the \irace{} entry point.
The execution starts from the \irace{} \texttt{run} command on the left, goes through the \ParadisEO{} modules, which call the \IOH{experimenter} problem (in blue) and loggers (in green).
After the run of the algorithm, a statistic is computed on the ECDF data (in cyan), which is then returned to \irace{} as performance metric (\emph{i.e.,} this is the ``fitness value'' that the evaluation associates to the configuration under evaluation).
Involved classes are represented using the UML convention.
For the sake of clarity, the \IOH{profiler} prefix is written as \texttt{IOH} and the type of the \texttt{eoAlgoFoundryFastGA} slots are indicated as \texttt{\{double\}} instead of \texttt{eoOperatorFoundry<double>}.}
\label{fig:architecture}
\end{figure}

\section{The Modular Benchmark Pipeline}
\label{sec:pipeline}
 Figure~\ref{fig:architecture} summarizes our automated algorithm design pipeline for the concrete use-case that will be studied in Sec.~\ref{sec:usecase}. The pipeline links 
 an algorithm configurator with an algorithm generator and a benchmark platform. The algorithm configurator asks the algorithm generator to instantiate an algorithm, which then solves a problem of the benchmark platform while being observed by a logger. After this run, the logger's data are summarized as a scalar performance measure, which is sent back to the algorithm configurator. We briefly present in this section the different components of our pipeline, and explain the reasons behind our choices. 

\paragraph{Algorithm Framework: \ParadisEO{}} Many evolutionary algorithms share similar design patterns, and are often composed of similar operators. This has given rise to several platforms which aim at supporting their users in designing evolutionary heuristics by compiling a set of readily-available operators within a standardized software environment. Given the substantial work that has been put into these frameworks, we decided to build our pipeline around one of the most powerful toolboxes.
To this end, we have ranked 39 frameworks among the ones easily available on the web, based on an adhoc metric combining rapidity, activity, features and license, \emph{e.g.,}~\cite{jMetal15,ECJ1,OpenBeagle,Jenetics,ECF,DEAP,Cllib}, to name only a few. 
Since speed is a major concern for our pipeline, we favor frameworks written in C++. To select an up-to-date framework and to ensure availability of support in case of technical issues, we also checked the contribution activity in recent years.
These two criteria reduced our choices to \ParadisEO{}~\cite{ParadisEO,ParadisEOjournal}, OpenBeagle~\cite{OpenBeagle}, and ECF~\cite{ECF}. Among these three, \ParadisEO{} covers the largest portfolio of algorithm families, which are composed in the framework by assembling atomic functions (called {\em operators}). \ParadisEO{} is also the most actively maintained framework among the three, so that we decided to use it for our work. 

The upper part of Figure~\ref{fig:architecture} shows the core classes of \ParadisEO{} involved in our setting. 

\paragraph{Algorithm Configuration: \irace{}} Several algorithm configuration tools have been developed in the last decade. Among the most common ones used in our community are irace~\cite{irace}, SMAC~\cite{SMAC}, SPOT~\cite{SPOT}, GGA~\cite{GGA}, and hyperband~\cite{hyperband}. 
We have chosen \irace{}\footnote{Version \href{https://github.com/MLopez-Ibanez/irace/releases/tag/v3.4.1}{3.4.1} of \url{https://cran.r-project.org/web/packages/irace/}, ran with {\em R} 3.6.3.} for this study, for practical considerations (previous experience, availability of documentation, support from development team). 

\paragraph{Experimental Environment: \IOH{experimenter}} The \IOH{profiler} project~\cite{IOHprofiler} is a modular platform for algorithm benchmarking of iterative optimization heuristics (IOH).
Within this project, \IOH{expe\-ri\-men\-ter} provides synthetic benchmarks which are very fast to execute and a standardized way of observing algorithms behavior through so-called {\em loggers}.
We have chosen this platform, because it is fast and its modular design made it particularly easy for us to integrate the algorithm design framework (being written in C++, as \ParadisEO{}).
\IOH{profiler} is also actively maintained, and provides access to broad ranges of different optimization processes.

Compared to Nevergrad~\cite{nevergrad}, we particularly like the detailed logging options, which provide information about the anytime behavior of the algorithms---information that is currently not available in Nevergrad.
Compared to the COCO~\cite{hansen2016coco} environment, \IOH{profiler} makes it considerably easier to test algorithms' performance on our own benchmark problems or suites.
Finally, the project also supports interactive performance analysis and visualization module, \IOH{analyzer}, which we used for the interpretation of our data. 

The lower part of Figure~\ref{fig:architecture} shows the classes related to the loggers and the problems that are used in our experimental study in Sec.~\ref{sec:usecase}.

\paragraph{Data Records: fast ECDF Logger} In our use-case, we decide to tune algorithms for good anytime performance, and to use volume under the approximated empirical cumulative density function (ECDF) curve as objective. To this end, we implement within \IOH{experimenter} 
an efficient way of computing these values. This ``ECDF logger'' will be described in Sec.~\ref{sec:ECDFlogger}.

\paragraph{Data Analysis and Visualization with \IOH{analyzer}} Data analysis and visualization is performed via \IOH{analyzer}~\cite{IOHanalyzer}, another module of the \IOH{profiler} project~\cite{IOHprofiler}. 
 
%%%%%%%%%%%%%%%%%%%%%%%%%%%% 
\section{Use-Case and Experimental Setup}%: Tuning Anytime Performance of Genetic Algorithms on W-Model Problems}
\label{sec:usecase}

Our use-case is the optimization of the anytime performance of a genetic algorithm on selected instances of the W-model problem. Our performance measure (Sec.~\ref{sec:ECDFlogger}), the algorithmic framework (Sec.~\ref{sec:GA}), and the problems (Sec.~\ref{sec:Wmodel}) are introduced in the first three subsections. We then summarize the experimental setup of the whole pipeline in Sec.~\ref{sec:setup}.
Our objective is to find the best algorithm for each instance, which would be the first step of a per-instance, landscape-aware algorithm selection, for instance. We thus do not consider training versus test sets.

\subsection{Anytime Performance Measure: AUC}
\label{sec:ECDFlogger}

In order to allow for large scale experiments, we implement a fast logger within \IOH{experimenter}, which essentially stores a histogram of the two-dimensional distribution of the number of runs having reached a quality/time target.
The time dimension is given as the number of calls to the objective function, linearly discretized between zero and the allowed budget.
The quality dimension is given as the absolute value of the best solution found during the run, linearly discretized between zero and the known $V_{\max}$ bound (see Table~\ref{tab:functions}).
This is essentially a discrete version of the Empirical Attainment Function~\cite{DBLP:conf/emo/FonsecaFH01}, which is related itself to the multivariate Empirical Cumulative Distribution Function~\cite{GRUNERTDAFONSECA2002179}. 

\begin{figure}[t]%[htbp]
    \centering
    \includegraphics[width=0.49\columnwidth]{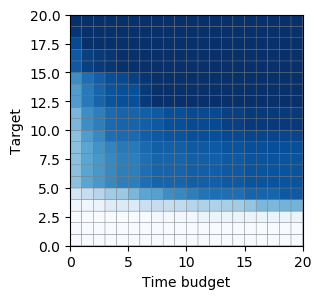}
    \hfill
    \includegraphics[width=0.49\columnwidth]{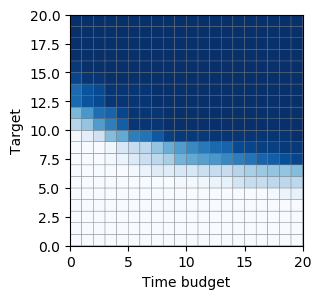}
    \caption{Example of two ECDF histograms. Abscissa shows the time dimension and the ordinate shows the target quality. The colormap shows the probability for the considered algorithm to reach a quality/time target falling in each bucket of the histogram. The figures show ECDF histograms for the 1ptGA algorithm (see Sec.~\ref{sec:GA}) after 50 independent runs on problem 2 (left) and of problem 5 (right), respectively, using 20 buckets on each axis.
    }
    \label{fig:ECDF_histogram}
\end{figure}

Figure~\ref{fig:ECDF_histogram} shows two examples of such histograms, arbitrarily chosen. The matrix defines the considered quality/time targets $(v,t)$. The color of each cell corresponds to the probability that the algorithm has identified, within the first $t$ function evaluations, a solution of quality at least $v$. The darker a cell, the larger the fraction of runs that could successfully meet the quality/time target. Note here that we assume minimization as objective. 

Using the histogram of the performance ECDF instead of its continuous counterpart allows to keep the data in-memory, in compact data structures, without having to rely on slow disk accesses.

The performance of the considered algorithm is computed as a statistic on this histogram. In our study, we use the volume under the curve (3D counterpart of the area under the curve, AUC) of the discretized ECDF, approximated as the sum of the EAF histogram. This allows for a compromise between quality and time, which is easily available because we consider synthetic benchmarks with known bounds.

\subsection{The $(\mu \stackrel{+}{,} \lambda)$ ``Fast'' GA Family }
\label{sec:GA}

\begin{algorithm2e}[t]
\textbf{Input:} 
Budget $B$, configuration $(\mu, \lambda, p_c, p_m)$, choice of the operators and conditional parameters. Note that $P$ and $P'$ are multi-sets, \textit{i.e.,} the same point may appear multiple times\;
\textbf{Initialization:}\\
\Indp
   $P \leftarrow$ InitialSampling$(\mu)$\;
   evaluate the $\mu$ points in $P$\;
   Evals $\leftarrow \mu$\;
\Indm		
	\textbf{Optimization:}
% 	\For{$t=1,2,3,\ldots$ {\normalfont{\textbf{until}}} $\Evals=B$}{
	\While{$\Evals<B$}{
	    $P' \leftarrow \emptyset$\;
	    \For{$i=1,\ldots,\lambda$}{
	        % Crossover with probability p_c
	        Sample $r_c \in [0,1]$ u.a.r.\;\label{line:SampleRc}
	        \eIf{$r_c \le p_c$\label{line:CrossoverIf}}{
	            $\left(y^{(i,1)}, y^{(i,2)} \right)$ $\leftarrow$ SelectC$(P)$\; \label{line:SelectC}
	            %$z^{(i,1)}$ $\leftarrow \text{Crossover}(y^{(i,1)}, y^{(i,2)})$\;
	            $\left(y'^{(i,1)},y'^{(i,2)}\right)$ $\leftarrow \text{Crossover}\left(y^{(i,1)}, y^{(i,2)}\right)$\;
	            Sample $z^{(i,1)} \in \left\{y'^{(i,1)},y'^{(i,2)}\right\}$ u.a.r. \;\label{line:Crossover}
	            Sample $r_m \in [0,1]$ u.a.r. \;
	            \eIf{$r_m \le p_m$\label{line:CrossMutIf}}{$z^{(i,2)} \leftarrow \text{Mutation}\left(z^{(i,1)}\right)$\;}{$z^{(i,2)} \leftarrow z^{(i,1)}$\;\label{line:CrossMutElse}}
	        }{$z^{(i,1)} \leftarrow$ SelectM$(P)$\;\label{line:SelectM}
	        $z^{(i,2)} \leftarrow \text{Mutation}\left(z^{(i,1)}\right)$\;\label{line:Mutation}}
	        Evaluate $z^{(i,2)}$\; 
	        Evals $\leftarrow$ Evals$+1$\;
	        $P' \leftarrow P'\cup \left\{z^{(i,2)}\right\}$\;
	   } %for i=1,...,lambda        
    $P\leftarrow$ Replace$(P,P',\mu)$\;\label{line:Replace}
	}
\caption{A Configurable Family of $(\mu \stackrel{+}{,} \lambda)$~Genetic Algorithms. }
\label{alg:GA}
\end{algorithm2e}

We chose for our use-case a family of $(\mu \stackrel{+}{,} \lambda)$ GAs, which is to a large extend inspired by the study~\cite{YeWDB20}. Algorithm~\ref{alg:GA} summarizes the framework, called ``FastGA'' in the implementation.

Essentially, given a parent population of $\mu$ points, each of the $\lambda$ offspring is created by first deciding which variation operator is applied (line~\ref{line:SampleRc}): with probability $p_c$ the offspring is generated by first recombining two search points from the parent population (lines \ref{line:SelectC}--\ref{line:Crossover}) and then randomly deciding (with probability $p_m$) whether or not to apply a mutation operator to the so-created offspring (lines \ref{line:CrossMutIf}--\ref{line:CrossMutElse}). When crossover was not selected in line~\ref{line:CrossoverIf}, the offspring is created by mutation (lines \ref{line:SelectM}--\ref{line:Mutation}). When all $\lambda$ offspring have been created, the iteration is completed by a replacement step (line~\ref{line:Replace}).

\paragraph{Implementation of this Family in \ParadisEO{}:} 

We implement this family of GAs through \ParadisEO{}'s ``foundries'', which allow to register a set of operators (\emph{e.g.,} several kind of mutations) within a ``slot'' (\emph{e.g.,} the step at which mutation is called within the algorithm). Before each call, it is possible to instantiate a specific operator among the registered ones, for each slot, thus assembling one of the algorithm instance among all the possible combinations of operators. Note that operators can be simple numbers, like a probability. Operators are referenced within slots by their indices.

Most of the operators we use were already available in \ParadisEO{}, to the exception of mutations operators with indices 1--5 (see below), which we implemented for this study. We also implemented the algorithm~\ref{alg:GA} as the \texttt{eoFastGA} class\footnote{All our code is contributed to the \ParadisEO{} project.}, in which to plug the operators.

We consider the following operators and parametrizations, which result in a (large) total number of 1\,630\,475 different configurations of Algorithm~\ref{alg:GA}. 
Numbers in brackets indicate the indices of the corresponding operators within its slot.

\paragraph{\textbf{InitialSampling$(\mu)$:} Initialization of the Algorithm (1 option)} We only consider independent uniform sampling, \textit{i.e.}, the $\mu$ points are i.i.d. uniform samples. The corresponding ParadisEO operator is \texttt{eoInitFixedLength}.

\paragraph{\textbf{Crossover rate $p_c$} (6 options).} 
We consider $p_c \in \{0, 0.2, 0.4, 0.5, 0.6, 0.8\}$. Being only able to use the {\em integer} and {\em categorical} interface for \irace{}. 

\paragraph{\textbf{SelectC$(P)$:} Selection of two points for the crossover operation (7 options).} Note that in the implementation, the selection operator (line~\ref{line:SelectC}) is called twice to select the two candidate points.
        \begin{description}
            \item [0] \texttt{eoRandomSelect()}:  Uniformly select a point from $P$ (without removing the first selected individual from the set $P$ used by the second selection). 
            (1 option).
            
            \item [1] \texttt{eoStochTournamentSelect($k$)}: Select a point from $P$ with tournament selection, \textit{i.e.}, we select uniformly at random  $k$ different points in $P$ and the best one of these is selected. $k$ denotes the tournament size as percentage of population (\textit{i.e.,  $k \in [0, 1]$}). (1 option,  $k = 0.5$). 
            
            \item [2] \texttt{eoSequentialSelect()}: Select the best point from $P$ (with respect to the objective function value). This operator is sometimes referred to as \emph{elitist selection} or \emph{truncation selection}. When called twice, it selects the two distinct best points from $P$. (1 option).

            \item [3] \texttt{eoProportionalSelect()}:  Select a point from $P$ with so-called fitness-proportional selection, \textit{i.e.}, point $x \in P$ is chosen with probability $f(x)/\sum_{y \in P}f(y)$. (1 option).
            
            \item [4--6] \texttt{eoDetTournamentSelect($k$)}: Like \texttt{eoDetTournamentSelect}, but $k$ is deterministic. (3~different options, each one for $k \in [2, 6, 10]$).
        \end{description}

\paragraph{\textbf{Crossover$(x,y)$:} Bivariate Variation Operators (11 options).}        
        \begin{description}
            \item [0--4] \texttt{eoUBitXover($b_c$)}: Uniform crossover with bias (or ``preference'' in ParadisEO) $b_c$, setting (independently for each position $i \in [1..n]$) $z_i = x_i$ with probability $b_c$ and setting $z_i = y_i$ otherwise. $z$ denotes the offspring element coming from the crossover of $x$ and $y$. (5 different options, $b_c \in [0.1, 0.3, 0.5, 0.7, 0.9]$).
            
            \item [5--9] \texttt{eoNPtsBitXover($k$)} : $k$-point crossover, which selects $i_1,\ldots, i_k$ uniformly at random and without replacement from $[1..n]$ and sets $z_i = x_i$ for $i \in [1..i_1] \cup [i_2+1 .. i_3] \cup \ldots$ and sets $z_i = y_i$ for $i \in [i_1+1..i_2] \cup [i_3+1 .. i_4] \cup \ldots$ (5 different options, $k \in [1, 3, 5, 7, 9]$). 
            
            \item [10] \texttt{eo1PtBitXover()}: Classic 1-point crossover. (1 option). We mistakenly added this option even if it is the same as the previous crossover with $k=1$. 
        \end{description} 
        
\paragraph{\textbf{Mutation probability $p_m$} (6 options):} We consider $p_m \in \{0, 0.2, 0.4, 0.5, 0.6, 0.8\}$.         
        
\paragraph{\textbf{Mutation$(x)$:} Univariate Variation Operator (11 options)} 
       All mutation operators are unary unbiased in the sense proposed in~\cite{LehreW12}. For a compact representation, we follow the characterization suggested in~\cite{DoerrDY20} and define the mutation operators via the distributions that they define over the possible mutation strengths $k \in [0..n]$. After sampling $k$ from the operator-specific distribution, the $k$-bit flip operator, $\flip_k(\cdot)$, is applied; it flips the entries in $k$ uniformly chosen, pairwise different bits (\textit{i.e.}, the $k$ bits are chosen u.a.r. without replacement).  
        \begin{description}
            
			\item [0] \texttt{eoUniformBitMutation()}: The ``uniform'' mutation operator, which samples $k$ uniformly at random in the set $[0..n]$. % and then applies flip$_{k}(\cdot)$.  
			(1 option).
			
            \item [1] \texttt{eoStandardBitMutation($p = 1/n$)}: This is the standard bit mutation with mutation rate $p$. It chooses $k$ from the binomial distribution $\mathcal{B}(n,p)$. % and then applies the flip$_{k}(\cdot)$ operator. 
            (1 option).
            
            \item [2] \texttt{eoConditionalBitMutation($p = 1/n$)}: A conditional standard bit mutation operator with mutation rate $p$. It chooses $k'$ from $\mathcal{B}(n-1,p)$ and applies the flip$_{k}(\cdot)$ operator with $k=k'+1$. (1 option).
            
            \item [3] \texttt{eoShiftedBitMutation($p = 1/n$)}: The ``shifted'' standard bit mutation with mutation rate $p$, suggested in \cite{CarvalhoD18arxiv}. It samples $k'$ from the binomial distribution $\mathcal{B}(n,p)$. When $k'=0$, it uses $k=1$ and it uses $k=k'$ otherwise. %It then applies the flip$_{k}(\cdot)$ operator. 
            (1 option).
            
            \item [4] \texttt{eoNormalBitMutation($p$,$\sigma^{2})$}: The ``normal'' mutation operator suggested in \cite{YeDB19}. It samples $k$ from the normal distribution $\mathcal{N}(pn,\sigma^2)$. When $k > n$, $k$ is replaced by a value chosen uniformly at random in the set $[0..n]$. 
            (1 option, $p = 1/n$ and $\sigma^{2} = 1.5$).
            
            \item [5] \texttt{eoFastBitMutation($\beta$)}: The ``fast'' mutation operator suggested in~\cite{fastGA}. It samples $k'$ from the power-law distribution $\mathcal{P}[L=k]=(C_{n/2}^{\beta})^{-1}k^{-\beta}$ with $C_{n/2}^{\beta}=\sum_{i=1}^{n/2} i^{-\beta}$. When $k'$ is larger than $n$, it samples a uniform value $k$ in $[0..n]$, and it uses $k = k'$ otherwise. 
            (1 option, $\beta = 1.5$).
            
            \item [6--10] \texttt{eoDetSingleBitFlip($k$)}: Deterministically applies $\flip_k(\cdot)$. (5 different options, $k \in [1, 3, 5, 7, 9]$).
            
            \end{description}
\paragraph{\textbf{SelectM$(P)$:} Selection of one point for the mutation operation if crossover was not chosen (7 options)} 
We essentially have the same selection operators as for crossover. The only difference is that we select only one point instead of two.

\paragraph{\textbf{Replace$(P,P',\mu)$:} Replacement of population (11 options)}
        \begin{description}
            \item [0] \texttt{eoPlusReplacement()}: The best $\mu$ points of the multiset $P\cup P'$ are chosen. (1 option).
            
			\item [1] \texttt{eoCommaReplacement()}: The best $\mu$ points of the offspring multiset $P'$ are chosen. (1 option).
            
            \item [2] \texttt{eoSSGAWorseReplacement()}: The $min(\lambda,\mu)$ points of the offspring multiset $P'$ replace the worst points in $P$. (1 option).
            
            \item [3--5] \texttt{eoSSGAStochTournamentReplacement($k$)}: Like\\\texttt{eoSSGADetTournamentReplacement}, $k$ being the the tournament size as percentage of population. (3 different options, $k \in [0.51, 0.71, 0.91]$).
			
            \item [6--10] \texttt{eoSSGADetTournamentReplacement($k$)}:  The $\mu$ points are selected through tournament selection. Each tournament involves $k$ uniformly chosen points in $P\cup P'$ and the best ones of these $k$ points is selected. This procedure is repeated $\mu$ times, each time removing an already selected point from the multi-set $P \cup P'$. (5 different options, $k \in [2, 4, 6, 8, 10]$).
        \end{description}

% Three extra operators define a configuration. In order to reduce the size of the search space a bit, we defined these three extra operators as constant ones (\textit{i.e.} the same for all the configurations) :
% \begin{itemize}
%     \item \texttt{aftercross\_selectors} : Define how to select beetween the two individuals altered by crossover which one will mutate. The only option for this operator is (\texttt{eoRandomSelect}).
%     \item \texttt{continuators} : Stopping criteria. We selected a generational continuator, the algorithm runs until a pre-defined budget \textit{nb\_gen} is reached (fixed to \textit{nb\_gen} = $\frac{B}{\lambda}$). The ParadisEO name for that operator is \texttt{eoGenContinue}. (1 component).
%     \item \texttt{offspring\_sizes} : The number of offsprings $\lambda$. 
% \end{itemize}        

This concludes our description of the high level operators of our family of $(\mu \stackrel{+}{,} \lambda)$ GAs. % our next target is to explain how a configuration is defined. 
The set of all combinations generates the algorithm design space on which we let \irace{} search for the configuration(s) that best solve a given problem instance.

% See Table~\ref{tab:configs} for an example with the results of the \irace{} configuration and the four baseline algorithms described next.  

\paragraph{Baseline Algorithms}
\label{sec:baselinealg}

We consider four baseline algorithms, against which we compare the results of the automated design. They were manually chosen without particular justification. 
% \begin{enumerate}
    % \item 
    \textbf{(1) $(\lambda+\lambda)$ EA:}  no crossover, plus replacement, standard bit mutation, random selector for mutations. % \johann{@Amine double check selector for mutation?}\carola{shall we just add the specification in Table~\ref{table:configs}?} \johann{If we lack space, we can.} \amine{The mutation selector is correct}
    % \item 
    \textbf{(2) $(\lambda+\lambda)$ fEA:} no crossover, plus replacement, fast bit mutation, random selector for mutations.  %\johann{@Amine double check selector for mutation}
    % \item 
    \textbf{(3) $(\lambda+\lambda)$ xGA:} sequential selections, uniform crossover, standard bit mutation, plus replacement, $p_c=0.4$, $b_c=0.4$. %\johann{@Amine @Carola, double check $p_c$ and $b_c$} \amine{Same here, the mutation selector is correct}
    % \item 
    \textbf{(4) $(\lambda+\lambda)$ 1ptGA:} sequential selections, 1-point crossover, standard bit mutation, plus replacement, $p_c=0.4$, $b_c=0.4$.
% \end{enumerate}
% \begin{tabular}{|*{5}{c|}}
%     \hline
%      \textbf{Configuration name} & \textbf{Description}\\
%     \hline
%      (5+5) EA & no crossover, plus replacement, standard bit mutation  \\
%      \hline
%      (5+5) fEA & no crossover, plus replacement, fast bit mutation\\
%     \hline
%      (5+5) xGA & seq selection, uniform crossover, standard bit mutation, plus replacement \\
%     \hline
%      (5+5) 1ptGA &  seq selection, 1-point crossover, standard bit mutation, plus replacement   \\
%     \hline
% \end{tabular}
% \captionof{table}{\textit{Benchmark configurations for the tests}}

%%%
\subsection{The W-Model Problems}
\label{sec:Wmodel}

We evaluate our automated algorithm design pipeline on the W-model functions originally suggested in~\cite{Wmodel}. In a nutshell, the W-model is a benchmark problem generator, which allows to tune different characteristics of the problems, see below for a description. We selected from this family of benchmark problems the 19 instances suggested in~\cite{WmodelInstancesASoC}, which are summarized in Table~\ref{tab:functions}. Note  here that the description differs from that given in~\cite{WmodelInstancesASoC}, since we used the implementation within \IOH{experimenter}, which was made available in the context of the work~\cite{DoerrYHWSB20}. 
The problem instances listed in Table~\ref{tab:functions} are identical to those suggested in~\cite{WmodelInstancesASoC}, it is only the representations that differ.  

It was suggested in~\cite{DoerrYHWSB20} to superpose the W-model transformations to different optimization problems. The instances selected in~\cite{WmodelInstancesASoC}, however, were only selected from transformations applied to the \onemax{} problem 
$\OM : \{0, 1\} ~ \overrightarrow{} ~ [0..n], x ~ \overrightarrow{} ~ \sum_{i=1}^{n} x_i.$
The \onemax{} problem has a very smooth and non-deceptive fitness landscape. Due to the well-known coupon collector effect \cite{CouponFlajolet}, it is relatively easy to make progress when the function values are small, and the probability to obtain an improving move decreases considerably with increasing function values. The complexity of the \onemax{} problem can be considerably increased through the following %four 
W-model transformations.

    \textbf{(1) Neutrality} $W(., \mu_W ,., .)$: The bit string $(x_1, ..., x_n)$  is reduced to a string $(y_1, ..., y_m)$  with $m := n / \mu_W$, where $\mu_W$ is a parameter of the transformation (we use the subscript $W$ to indicate  that these are parameters of the W-model problem generator). For each $i \in [m]$ the value of $y_i$ is the majority of the bit values in the size-$\mu$ substring $(x_{(i-1)\mu_W}, x_{(i-1)\mu_W+1},...,x_{i\mu_W})$ of $x$. That is, $y_i = 1$ if and only if there are at least $\mu_W / 2$ ones in this ``block''. When $n / \mu_W \notin  \mathbb{N}$ , the last bits of $x$ are copied to $y$.
    %
    % \item 
    
    \textbf{(2) Epistasis} $W(., . ,\nu_W, .)$: Epistasis introduces local perturbations to the bit strings. It first ``cuts'' the input string $(x_1, ..., x_n)$ into subsequent blocks of size~$\nu_W$. Using a permutation $e_{\nu_W} : \{0,1\}^{\nu_W} ~ \overrightarrow{}~ \{0,1\}^{\nu_W} $, each substring $(x_{(i-1)\nu_W + 1}, x_{(i-1)\nu_W +2},...,x_{i\nu_W})$  is mapped to another string $(y_{(i-1)\nu_W+1}, y_{(i-1)\nu_W+2},...,y_{i\nu_W}) = e_{\nu_W}((x_{(i-1)\nu_W+1}, x_{(i-1)\nu_W+2},...,x_{i\nu_W}))$. The permutation $e_{\nu_W}$ is chosen in a way that Hamming-1 neighbors are mapped to strings of Hamming distance at least $\nu_W - 1$, see~\cite{Wmodel} for examples.  
    %
    % \item 
    
    \textbf{(3) Ruggedness and Deceptiveness} $W(., . ,., \gamma_W)$: This layer perturbs the fitness values, by applying a permutation $\sigma(\gamma_W)$ to the possible fitness values $[0..n]$. The parameter $\gamma_W$ can be thought of as a parameter which controls the distance of the permutation to the identity. The permutations $\sigma(\gamma_W)$ are chosen in a way such that the ``hardness'' of the instances monotonically increases with increasing $\gamma_W$, see~\cite{Wmodel} for details. 
% \end{enumerate}

We convert these functions into a minimization problem by multiplying all values by $-1$. 

\begin{table}[t]
 \caption{Test problems on which the pipeline is evaluated, taken from~\cite{WmodelInstancesASoC}. In column ``best'' we list the baseline algorithm with largest average AUC value, reported in column AUC$_{\text{b}}$. AUC$_{\text{i}}$ is the  average of the 15 mean AUC of elite configurations, as suggested by 15 independent runs of \irace{} with default hyperparameters. 
    AUC-values are w.r.t. to at least 50 validation runs and 
    ``rel.'' indicates the relative gain $(\AUC_{\text{i}}-\AUC_{\text{b}})/\AUC_{\text{b}}$.
    }
    \label{tab:functions}
    \centering
\begin{tabular}{l|lllll|lllr}
FID & dim & $\mu_W$ & $\nu_W$ & $\gamma_W$ & $V_{\max}$ & best & AUC$_{\text{b}}$ & AUC$_{\text{i}}$ & rel.\\
\hline                                         
1	&	20	&	2	&	6	&	10	&	10	&	xGA	&	8378	&	8740	&	4\%	\\
2	&	20	&	2	&	6	&	18	&	10	&	fEA	&	8402	&	8754	&	4\%	\\
3	&	16	&	1	&	5	&	72	&	16	&	fEA	&	8352	&	8397	&	1\%	\\
4	&	48	&	3	&	9	&	72	&	16	&	EA	&	8299	&	8914	&	7\%	\\
5	&	25	&	1	&	23	&	90	&	25	&	fEA	&	8003	&	8510	&	6\%	\\
6	&	32	&	1	&	2	&	397	&	32	&	1pt	&	7055	&	7311	&	4\%	\\
7	&	128	&	4	&	11	&	0	&	32	&	1pt	&	6833	&	8183	&	20\%	\\
8	&	128	&	4	&	14	&	0	&	32	&	EA	&	6885	&	8499	&	23\%	\\
9	&	128	&	4	&	8	&	128	&	32	&	xGA	&	8154	&	8786	&	8\%	\\
10	&	50	&	1	&	36	&	245	&	50	&	fEA	&	7216	&	8122	&	13\%	\\
11	&	100	&	2	&	21	&	256	&	50	&	EA	&	8314	&	9139	&	10\%	\\
12	&	150	&	3	&	16	&	613	&	50	&	EA	&	8034	&	8730	&	9\%	\\
13	&	128	&	2	&	32	&	256	&	64	&	fEA	&	8076	&	9345	&	16\%	\\
14	&	192	&	3	&	21	&	16	&	64	&	fEA	&	6173	&	7677	&	24\%	\\
15	&	192	&	3	&	21	&	256	&	64	&	fEA	&	6797	&	8292	&	22\%	\\
16	&	192	&	3	&	21	&	403	&	64	&	fEA	&	7273	&	8592	&	18\%	\\
17	&	256	&	4	&	52	&	2	&	64	&	xGA	&	6935	&	9028	&	30\%	\\
18	&	75	&	1	&	60	&	16	&	75	&	EA	&	5958	&	7089	&	19\%	\\
19	&	150	&	2	&	32	&	4	&	75	&	EA	&	7399	&	8717	&	18\%	\\
\hline
\end{tabular}
\end{table}

\subsection{Experimental Setup}
\label{sec:setup}

Our test bed is the automated design of Algorithm~\ref{alg:GA} with the options specified in Sec.~\ref{sec:GA} and with the objective to maximize the AUC as defined in Sec.~\ref{sec:ECDFlogger}, and this for each of the 19 problems listed in Table~\ref{tab:functions}. 
These instances of the W-model problem were suggested in~\cite{WmodelInstancesASoC} based on an empirical study using clustering of algorithm performance data, with the goal to select a diverse collection of benchmark problems. Note here that we tune the algorithms for each problem individually.
That is, we apply our algorithm design pipeline 19 independent times. 

For the sake of simplicity, we fix the population sizes to $\lambda=\mu=5$, for the search performed by \irace{} and for our baseline algorithms. 

For each use-case, we set the budget of the algorithms to $5n$ function evaluations (FEs). To compute the AUC, we evaluate the performance at 100 linearly distributed budgets $b^1, \ldots, b^{100} \in [1,5n]$ and at 100 linearly distributed target values $v^1, \ldots, v^{100} \in [0,V_{\max}]$. 
Linearization computes the bucket index $i = \lfloor (x - x_{\min}) / (x_{\max}-x_{\min}) \cdot 100 \rfloor$ for both budgets and targets. 

To find the best algorithm design, we allow \irace{} a budget of 100\,000 target runs.
We ensure that \irace{} performed at least 50 independent runs for the first ranked elite configuration,
adding additional runs if needed, and keeping all runs if \irace{} conducted more than 50 runs.
We do not consider the other elite configurations ranked by \irace{}, even in draw cases. 
We run \irace{} 15 independent times, to check the robustness of its selection.
We compare performance to the four baseline algorithms, which we run 50 independent times each on each of the 19 test problems. 

In total, these experiments took around $15\times 3$ hours 
on a computer with four Intel CPU cores i5-7300HQ at 2.50GHz and Crucial P1 solid-state disks. 

\section{Experimental Results}
\label{sec:results}

\paragraph{Comparison of AUC Values by Function} 
% \textbf{Comparison of AUC Values by Function.} 
Table~\ref{tab:functions} compares the AUC values of the best out of the four baseline algorithms against that of the elite configuration suggested by \irace{}. We observe that, for each of the 19 functions, the elite configurations suggested by \irace{} perform better than the best baseline algorithm. We report in Table~\ref{tab:functions} the average values, but the differences between the individual irace runs are very small, less than 2.1\% difference in AUC value between the best and the worst elite configuration for all 19 problems, and less than 1\% performance difference for 9 out of the 19 functions. 
The relative advantage of the \irace{} recommendations over the best baseline algorithms varies between 1\% and 30\%. When looking at each of the 15 elite configurations suggested per function, the best relative advantage is 31\% for F17, whereas two of the \irace{} elites performed worse than the best of the four baseline algorithms on function F3. For all other functions, all 15 \irace{}  elites have a better AUC value than the best of the four baseline algorithms. 
% even though we compare the {\em average} elite performance across the 16 runs. We see a clear advantage of the \irace{} configurations. 
However, although we see a clear advantage of the \irace{} configurations, we should keep in mind that the irace configurations are specifically tuned for each function, whereas the configurations of the four baseline algorithms are identical for all 19 W-model functions.  

Unfortunately, our pipeline does not yet allow to tune a single best solver, \textit{i.e.}, a single configuration that maximizes the AUC under the aggregated ECDF curve. Adding this functionality is a straightforward extension of our framework, which we plan to address in future work. The key challenge here is that \ParadisEO{} does not have the feature to easily reset on the fly the states of solvers between two runs on different problems.

\begin{table*}%[]
\caption{Configuration of the best out of the elite recommendation suggested by 15 independent runs of \irace{}, for each of the 19 benchmark problems specified in Table~\ref{tab:functions}, and compared against the configuration of the four baseline algorithms. The ``op.'' column gives the number of options per operator. All other integer values correspond to the indices with which the different options are listed in Sec.~\ref{sec:GA} and ``-'' indicates a non-applicable element (\emph{e.g.,} no crossover operator is used when $p_c=0$).}
\label{tab:configs}
{\footnotesize
\resizebox*{\textwidth}{!}{% <------ Don't forget this %
\begin{tabular}{lr|rrrrrrrrrrrrrrrrrrr|rrrr}
\toprule
\textbf{Operator}  & \textbf{op.} & 1 & 2 & 3 & 4 & 5 & 6 & 7 & 8 & 9 & 10 & 11 & 12 & 13 & 14 & 15 & 16 & 17 & 18 & 19 & EA & fEA & xGA & 1ptGA \\
\midrule
$p_c$              & 5            & 1 & 4 & 1 & 2 & 4 & 0 & 3 & 0 & 2 &  4 &  3 &  2 &  3 &  1 &  2 &  2 &  3 &  4 & 4  &  0 &  0  &  2  &    2  \\
\arrayrulecolor{lightgray}\hline
\textbf{SelectC}   & 7            & 2 & 5 & 3 & 1 & 2 & - & 0 & - & 2 &  2 &  2 &  2 &  6 &  5 &  5 &  2 &  2 &  2 & 2  &  - &  -  &  2  &    2 \\
\arrayrulecolor{lightgray}\hline
\textbf{Crossover} & 11           & 1 & 2 & 8 & 1 & 2 & - & 3 & - & 2 &  2 & 10 &  5 &  2 &  9 &  2 & 10 &  2 &  2 & 2  &  - &  -  &  2  &    5 \\
\arrayrulecolor{lightgray}\hline
$p_m$              & 5            & 2 & 3 & 2 & 2 & 4 & - & 4 & - & 4 &  4 &  4 &  4 &  4 &  4 &  4 &  4 &  4 &  4 & 4  &  - &  -  &  2  &    2 \\
\arrayrulecolor{lightgray}\hline
\textbf{SelectM}   & 7            & 2 & 4 & 6 & 6 & 3 & 2 & 3 & 2 & 5 &  5 &  2 &  3 &  1 &  2 &  6 &  6 &  5 &  1 & 6  &  0 &  0  &  2  &    2 \\
\arrayrulecolor{lightgray}\hline
\textbf{Mutation}  & 11           & 8 & 9 & 3 & 9 & 7 & 6 &10 &10 &10 &  9 & 10 &  9 & 10 &  8 &  8 & 10 & 10 &  8 & 9  &  1 &  5  &  1  &    1 \\
\arrayrulecolor{lightgray}\hline
\textbf{Replace}   & 11           & 8 & 9 & 2 & 0 & 0 & 0 & 0 & 0 & 0 &  0 &  0 &  0 &  0 &  0 &  0 &  0 &  0 &  0 & 0  &  0 &  0  &  0  &    0 \\
\arrayrulecolor{black}\bottomrule
\end{tabular}% <------ Don't forget this %
}
}%small
\end{table*}

\begin{table*}
\caption{Distribution of operators variants recommended by 15 runs of \irace{}, for problems 5 (left), problem 17 (center) and all problems (right).
The most selected indices are highlighted in bold and the darker the background color, the more often the operator instance is selected.
Empty cells indicates that \irace{} never selected the operator instance, cells with a ``-'' entry marks indices which are not defined for this operator.}
\label{tab:operators-distrib}
\centering
\includegraphics[width=\linewidth]{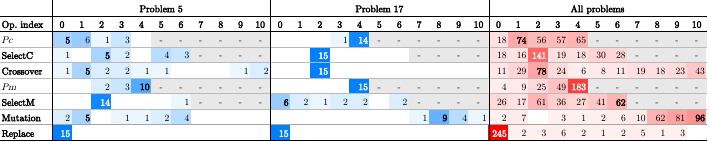}
\end{table*}

\paragraph{Comparison of the Configurations.} 
% \textbf{Comparison of the Configurations.} 
Table~\ref{tab:configs} summarizes the best of the 15 elite configurations that were suggested by \irace{} and compares them against the four baseline algorithms.
Table~\ref{tab:operators-distrib} shows the distribution of operators chosen by the 15 \irace{} runs. For the latter, we have chosen problem~17 as an example because we observed here the largest relative gain (see Table~\ref{tab:functions}). We have added problem~5 for comparison, because the distribution of operators suggested for it is very distinct from that of problem~17.

It is worth noting that each operator is selected at least once in the $19\times 15$ elite configurations suggested by \irace{} (Table~\ref{tab:operators-distrib}, right), which seems to confirm that i) different operators work well on different problems and ii) that \irace{} searches the full design space, giving some indication that it is not too large or too complex for automated tuning approaches.

We can see that, among all the best configurations proposed by \irace{} across 15 runs, none are similar to one of the baseline algorithms.
The probability of mutation $p_m$ is most frequently set to higher values and the most often chosen mutation is deterministic bit flip with larger number of bits (index~10 in the {\em Mutation} slot, which is the $\flip_{9}$ mutation operator). This indicates that larger mutation strengths could have been worth investigating, a result that has surprised us, since in most benchmark studies we see small mutation rates as defaults (albeit we do not consider various population sizes in this study).
The results confirm the superiority of the {\it plus} replacement (id.~0 in the {\em Replace} slot) and support the use of an elitist selection for the crossover candidates (id.~2 in {\em SelectC}).
We can also see that the uniform crossover with $b_c=0.5$ (id.~2 in {\em Crossover}) is more often chosen, like a small probability of performing a crossover (id.~1 in $p_c$).

For some problems, \irace{} almost always suggest a similar algorithm. On problem~17, for example, it often selects a GA with a large probability of applying uniform crossover in combination with deterministic bit flip mutations. 
For some other problems, a larger variance on the selected operators can be observed. For instance on problem~5, \irace{} selects a high mutation probability along with an elitist mutation selection, but does not show a clear preference for the other slots.

These results support the idea that there is not always a single best solver (\emph{i.e.,} ``No Free Lunch''), even when considering limited design and benchmarking spaces. We also see that some problems seem to require certain design choices, whereas others can be solved well by a broad range of configurations. A more detailed analysis of how these preferences correlate with the characteristics of the problems should offer plenty of interesting insights, but is left for future work.

\begin{figure}[t]
    \centering
    \includegraphics[width=1.0\columnwidth]{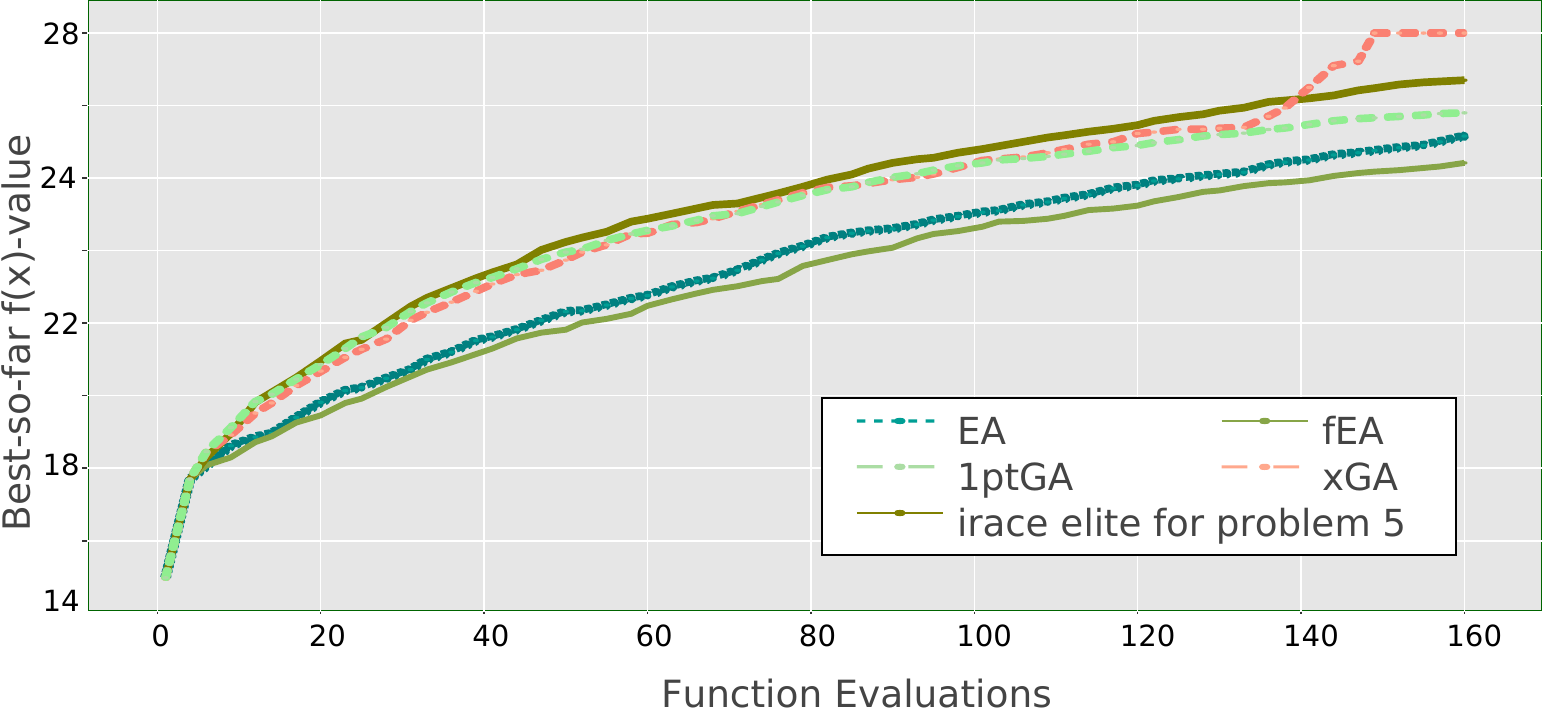} \\
    \vspace{0.5cm}
    \includegraphics[width=1.0\columnwidth]{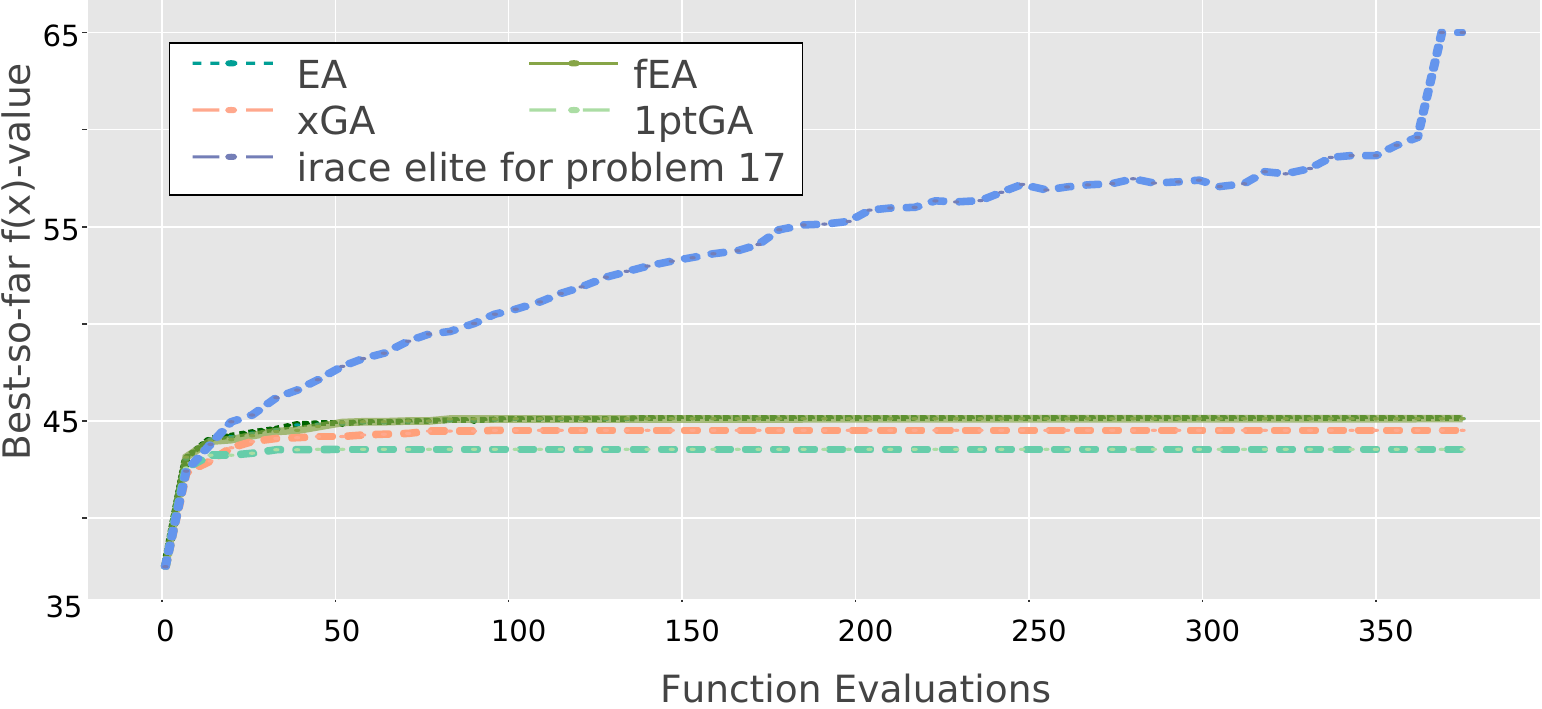}
    \caption{Convergence plots for the baseline algorithms and the elite configurations suggested by \irace{}, on problem~5 (top) and problem~17 (bottom).}
    \label{fig:bests-budgets}
\end{figure}

\paragraph{Fixed-budget solution qualities} 
% \textbf{Fixed-budget solution qualities.} 
Figure~\ref{fig:bests-budgets} shows two examples of convergence plots, where we plot the values of the best solutions found so far against the number of objective function evaluations performed, for each baseline algorithm and for the best elite configuration selected by \irace{}. Problems~5 and~17 are chosen to allow for comparison with Table~\ref{tab:configs}.

We observe that the elite configuration on problem~17 is largely more efficient than any of the baseline algorithms.
However, on problem~5, the elite configuration is only the most efficient until 140 evaluations.
It is selected nonetheless, because we consider the AUC of the 2D ECDF, which takes into account the average performance (across all budgets and targets) rather than the terminal budget of the best target. 
We believe that, whatever the performance metric we choose, there will always exists such artifacts, where some algorithm would be the best, had we chosen another metric.
It is clear, however, that even in this plot the elite configuration performs better {\em most of the time}.

\section{Conclusions and Future Work}
\label{sec:conclusions}

By interfacing the three state-of-the-art benchmarking modules from the evolutionary computation literature, \irace{}~\cite{irace}, \ParadisEO{}~\cite{ParadisEO}, and \IOH{profiler}~\cite{IOHprofiler}, we have introduced in this work a powerful pipeline for the automated design of sampling-based optimization algorithms. We have demonstrated its efficiency on the use-case of tuning a family of genetic algorithms on instances of the W-model~\cite{Wmodel} suggested in~\cite{WmodelInstancesASoC}.

Our results supports the idea that automated algorithm design can lead to increased performances and that there are efficient designs which wait to be studied more thoroughly.
We believe that efficient pipelines like the one we introduce has the potential to help raising the level of abstraction at which researchers are working.
Using automated algorithm design, it becomes possible to check if a newly designed operator can actually be useful in some algorithms/problems coupling~\cite{XuHHL12}.
We also believe that such studies can help deriving generic rules about algorithm design and could probably help theoretical researchers by suggesting {\em where} to look for interesting structures.

The modular design of the pipeline and its components makes our approach very broadly applicable. It is not restricted to particular types of problems nor to specific algorithms. In particular, extensions to continuous or mixed-integer problems are rather straightforward. Indeed, the \ParadisEO{} framework is designed to separate operators which are independent of the encoding (selection, replacement, etc.) from operator which depends on it (mutation, crossover, etc.), allowing for easy reuse of components and extensions to other algorithmic paradigms (estimation of distribution, local search, multi-objective, etc.). Additionally, the \IOH{experimenter} provides loggers for vectorial encodings and benchmarks for both numerical and bitstring encodings.

Our work is partially motivated by an industrial application that requires an automated configuration of hardware products. However, we believe that our pipeline is not only interesting for such practical purposes. For researchers, our pipeline offers an elegant way of assessing new algorithm operators and their interplay with already existing ones. 

In terms of further development, we plan to add the necessary features which would i) allow for running the same algorithm on multiple problems, while using a single logger that aggregates the results and would ii) support \irace{}'s interface for numerical parameters (additionally to categorical and integer ones). 

We then plan to test the approach on different algorithms families, with a possible extension to generic ``bottom-up'' hybridization grammars~\cite{marmion2013autodesign} and studies on the most efficient algorithms design (\emph{e.g.,} on the correlations between elite algorithms' operators).

We also plan to extend the framework by integrating feature extraction methods that use algorithm trajectory data~\cite{DerbelLVAT19,BajerPRH19} and/or samples specifically made for exploratory landscape analysis~\cite{mersmann2011exploratory,flacco} to couple the algorithm design to such information, similar to the per-instance configuration approaches made in~\cite{HutterHHL06PIAC,BelkhirDSS17}.

Our long-term vision is a pipeline for the automated design of algorithms which adjust their behavior during the optimization process, by taking into account information accumulated so far, similar to the dynamic algorithm configurations studied under the notion of \emph{parameter control}~\cite{KarafotiasHE15}. 
In contrast to the static designs considered in this work, the automated design of dynamic algorithms requires to select suitable update rules (e.g. based on time, on progress, on self-adaption, etc.). 

Finally, we also consider interesting the idea to provide a user-friendly front-end which allows users to assemble a benchmark study by selecting (e.g. through a graphical user interface) one or more algorithms and problems, the budget, etc. and then passing on this study to an automated interface which tunes (if desired) and runs the algorithm(s) and then automatically directs its users to the data summary and visualization platform \IOH{analyzer}, where the results of the empirical study can be analyzed. We believe that such a pipeline would greatly improve the deployment of evolutionary methods in practice.

\vspace{1.5ex} {{%{\small{
\textbf{Acknowledgments.} 
We thank the GECCO ECADA workshop reviewers for very constructive feedback and for pointers to~\cite{DBLP:journals/eor/Lopez-IbanezS14}. 

 This work has been financially supported by the Paris Ile-de-France region. 
%This work was partially done during the M2 Master internship of Amine Aziz-Alaoui at \'Ecole Polytechnique, Institut Polytechnique de Paris, CNRS, LIX, Palaiseau, France.
 }}

%\bibliographystyle{amsalpha}
%\bibliography{referencestwo}
\newcommand{\etalchar}[1]{$^{#1}$}
\providecommand{\bysame}{\leavevmode\hbox to3em{\hrulefill}\thinspace}
\providecommand{\MR}{\relax\ifhmode\unskip\space\fi MR }
\providecommand{\MRhref}[2]{%
  \href{http://www.ams.org/mathscinet-getitem?mr=#1}{#2}
}
\providecommand{\href}[2]{#2}

\appendix 

%The following pages forms the supplementary material to the ``Large Scale Algorithm Design'' article.

\newpage
\section{Comparison of frameworks}
\label{sec:comparison-frameworks}

\begin{table}[h!]
\caption{Comparison of software frameworks for evolutionary computation. Rank is based on an adhoc aggregation of subjective metrics based on performance of the programming language, activity of the project (number of contributors), breadth of features (number of modules, number of lines of code), and ease of integration in industrial projects (license). Data has been gathered in 2019.}
\centering
\includegraphics[width=0.99\textwidth]{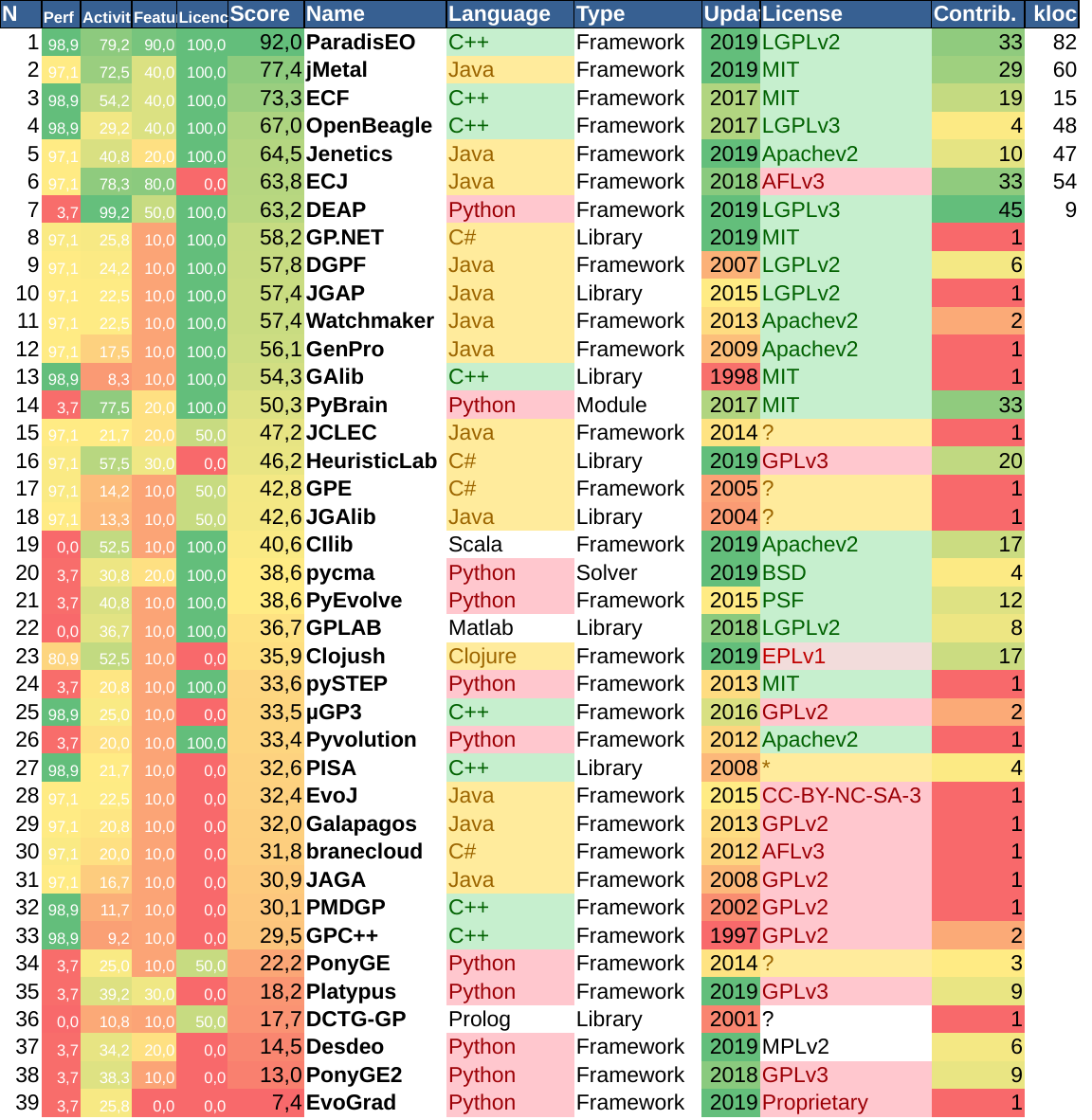}
\label{tab:comparison-frameworks}
\end{table}

\newpage
\section{Average AUC values}
\label{sec:average-auc}

\begin{table}[h!]
    \caption{Average AUC values of the elites returned by irace in each of the 15 independent runs and of the four baseline algorithms on each of the 19 W-model instances}
    \label{fig:my_label}
    \centering %trim=left bottom right top, clip
    \includegraphics[width=\linewidth, trim=1.7cm 37.5cm 2cm 2cm, clip]{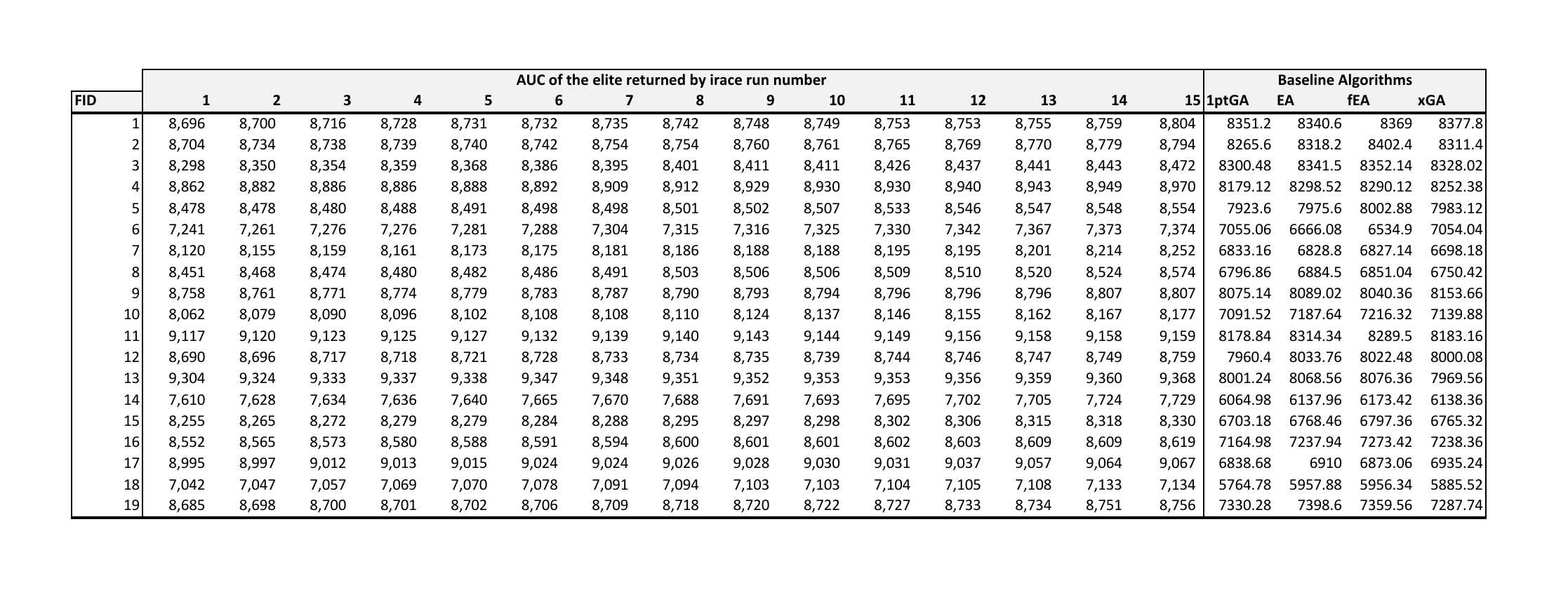}
\end{table}

\begin{figure}[h!]
    \centering
    \includegraphics[width=0.7\textwidth]{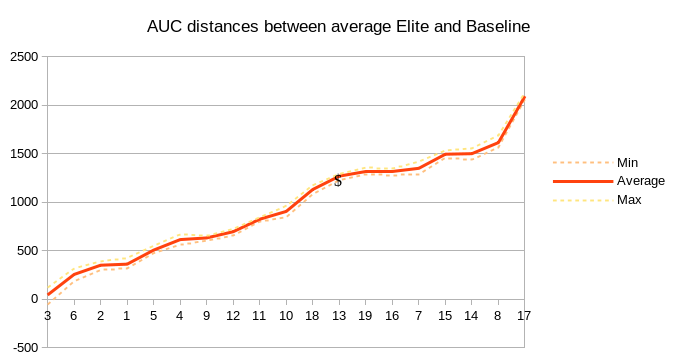}
    \caption{Distances between AUCs of elite algorithms and baseline algorithms.}
    \label{fig:AUC-distances}
\end{figure}

\newpage
\section{Diagram of \ParadisEO{} classes}
\label{sec:paradiseo-classes}

\begin{figure*}[h!]
\label{fig:grammar}
\includegraphics[width=1.0\textwidth]{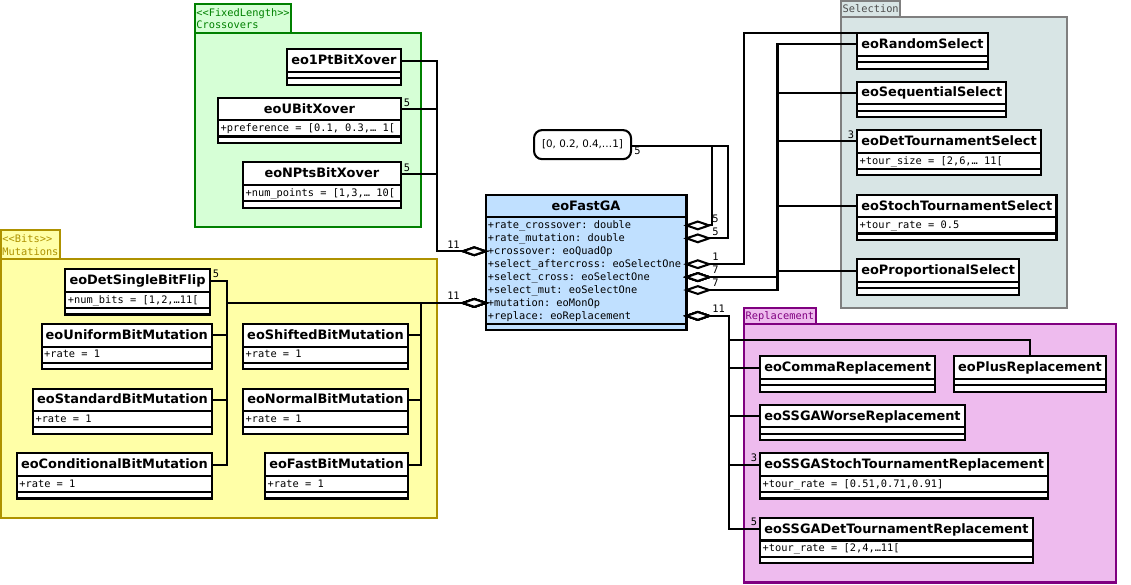}
\caption{Summary UML diagram of the FastGA family of algorithms, as modeled with \ParadisEO{} classes. Aggregation arrows shows the cardinality of instances (arrow tail side) and slots (arrow head side) involved in the final combination. No cardinality is indicated when it equals one.
%\johann{Double check consistency with the text of the article for both names and cardinalities.} % \johann{done}
%\johann{This essentially summarize the text, may be removed without loss if in need of space.}\carola{I suggest we move this to the appendix, as we will be struggling with space indeed}
}
\end{figure*}

% \section{Configuration Space for Algorithm~1}

% \begin{table}[h!]
%     \centering
% \begin{tabular}{|*{4}{c|}}
%     \hline
%      \textbf{Operator name} & \textbf{Description} & \textbf{Nb of Options} \\
%     \hline
%     \texttt{crossover\_rates}	& $p_c$, the crossover probability &	6 : $p_c \in [0, 0.2, 0.4, 0.5, 0.6, 0.8]$ \\
%      \hline
%     \texttt{crossover\_selectors} &	\textbf{SelectC} operators defined in 3.2	& 7 options\\
%     \hline
%     \texttt{crossovers} & \textbf{Crossover} operators defined in 3.2 & 11 options \\
%     \hline
%     \texttt{mutation\_rates} & $p_m$, the mutation probability & 6 : $p_m \in [0, 0.2, 0.4, 0.5, 0.6, 0.8]$ \\
%     \hline
%     \texttt{mutation\_selectors} & \textbf{SelectM} operators defined in 3.2 & 7 options \\
%     \hline
%     \texttt{mutations} & \textbf{Mutation} operators defined in 3.2 & 11 options \\
%     \hline
%     \texttt{replacements} & \textbf{Replace} operators defined in 3.2 & 11 options \\
%     \hline
% \end{tabular}
% \caption{Description of the configuration space on which we apply \irace{} to optimize the design of Algorithm~\ref{alg:GA} }
%     \label{tab:config-space}
% \end{table}

\newpage
\section{Convergence plots for all problems}
\label{sec:convergence-plots}

The following 19 figures shows the convergence plots of the baseline algorithms against the best elite selected by \irace{}.
Algorithms are denoted in the legend by the set of indices for each slots, using the following code (see Table~\ref{tab:configs} for the corresponding algorithms):
\begin{enumerate}
    \item [P:] population size (always 5 in this study),
    \item [C:] crossover probability,
    \item [s:] crossover selector,
    \item [c:] crossover,
    \item [a:] selector after crossover (always 0 in this study),
    \item [M:] mutation probability,
    \item [u:] mutation selector,
    \item [m:] mutation,
    \item [r:] replacement,
    \item [O:] stopping criterion (always 0 in this study).
\end{enumerate}

\begin{figure}
    \centering
    \includegraphics[width=1.0\textwidth]{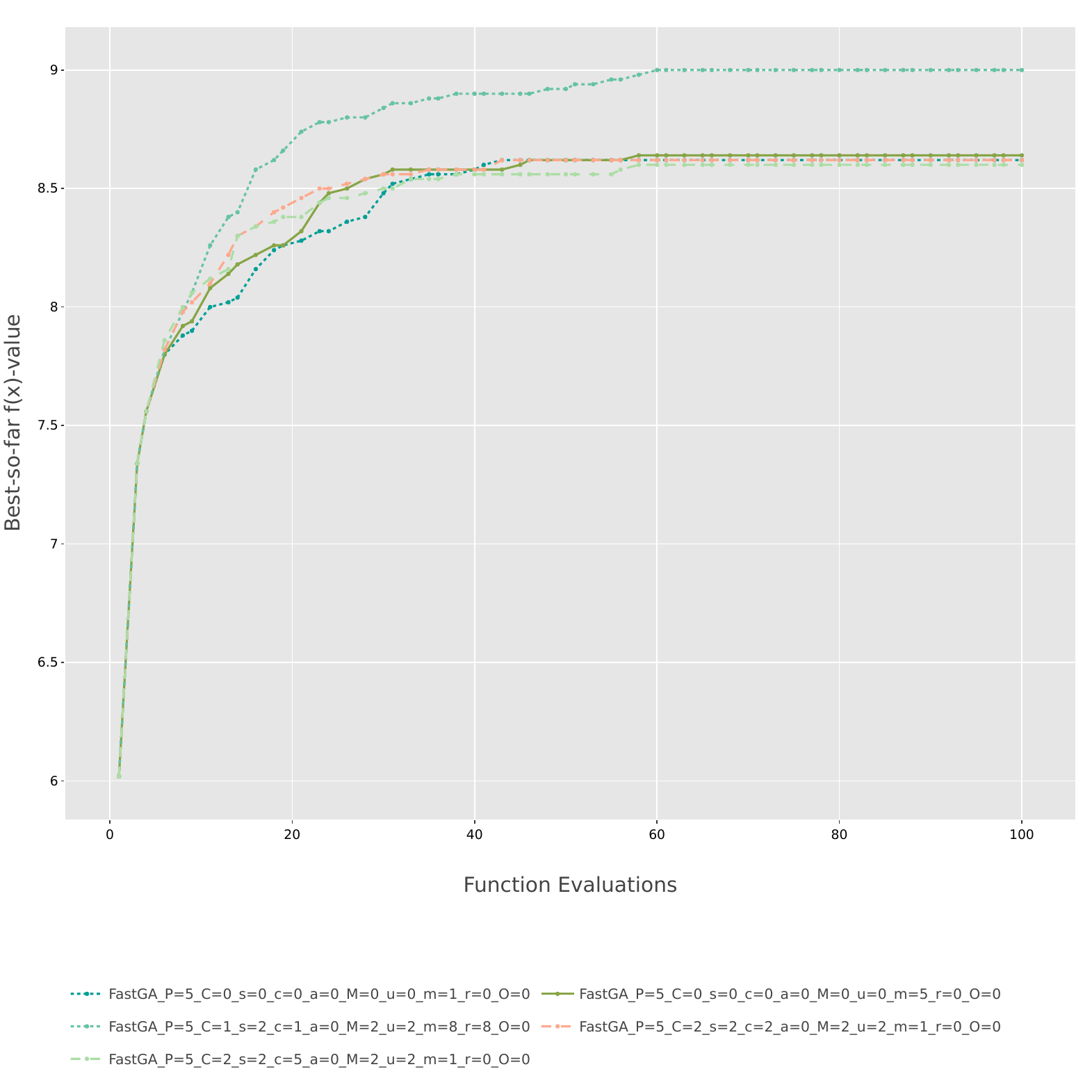}
    \caption{Convergence plot of baseline algorithms and elite, for problem 1.}
    \label{fig:convergence-1}
\end{figure}

\begin{figure}
    \centering
    \includegraphics[width=1.0\textwidth]{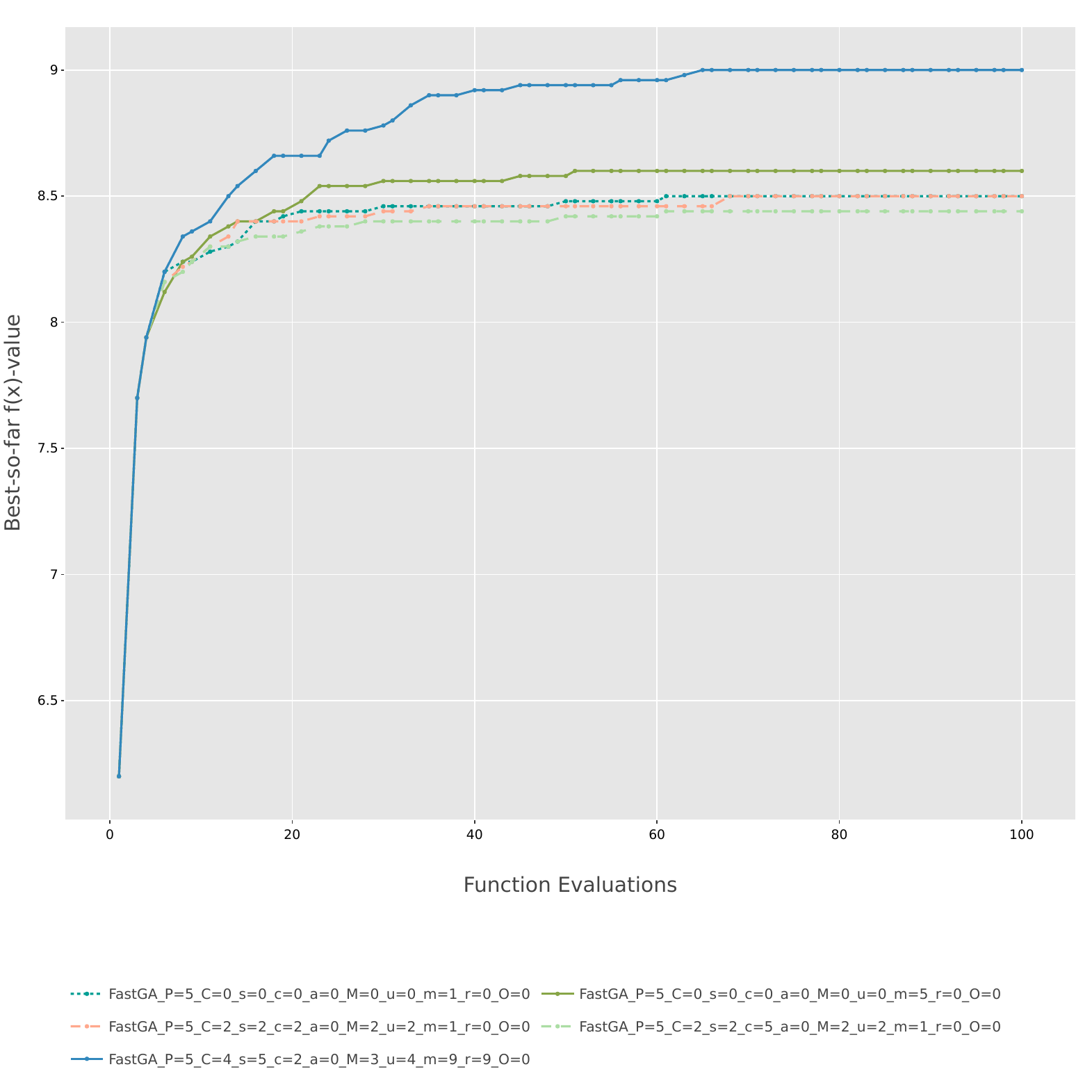}
    \caption{Convergence plot of baseline algorithms and elite, for problem 2.}
    \label{fig:convergence-2}
\end{figure}

\begin{figure}
    \centering
    \includegraphics[width=1.0\textwidth]{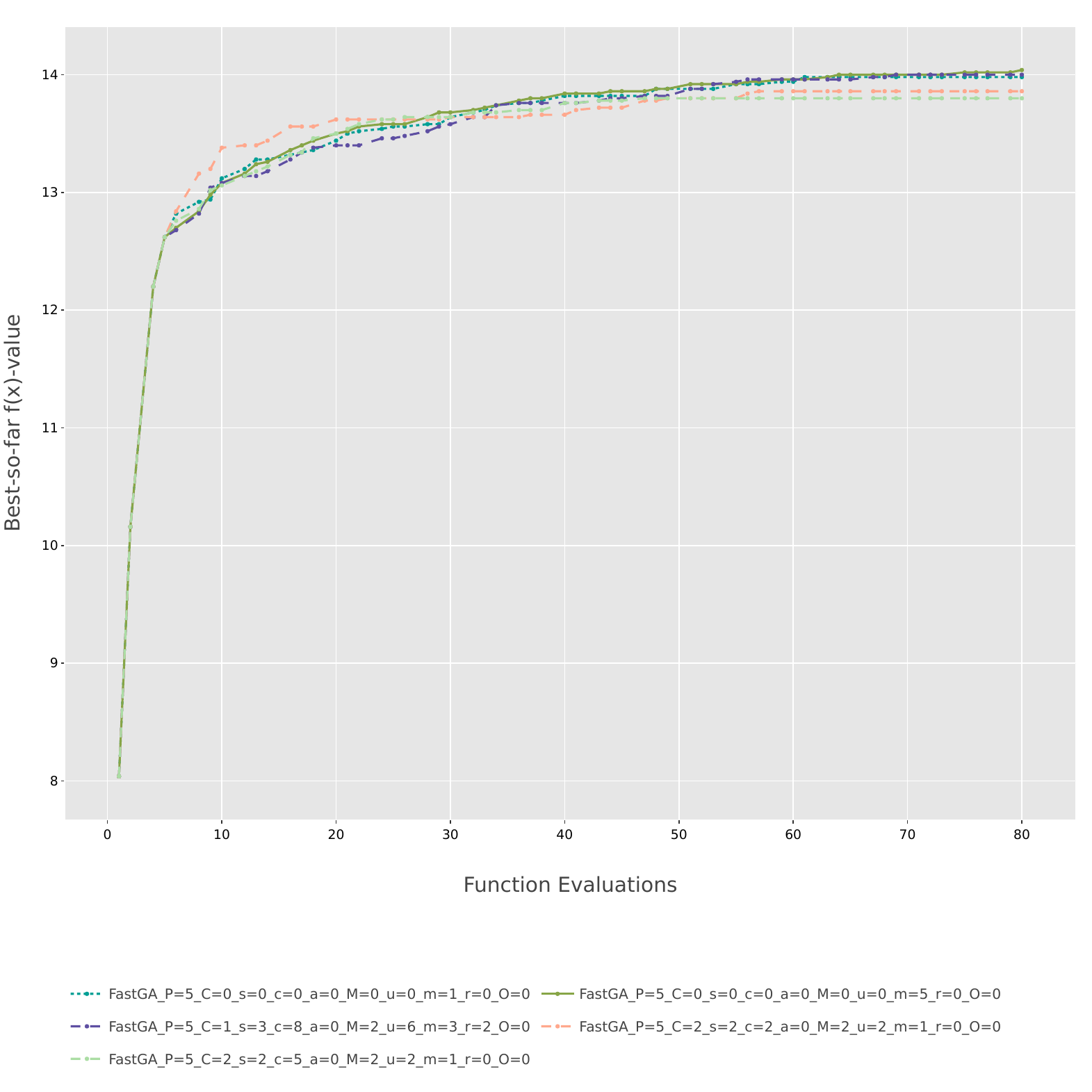}
    \caption{Convergence plot of baseline algorithms and elite, for problem 3.}
    \label{fig:convergence-3}
\end{figure}

\begin{figure}
    \centering
    \includegraphics[width=1.0\textwidth]{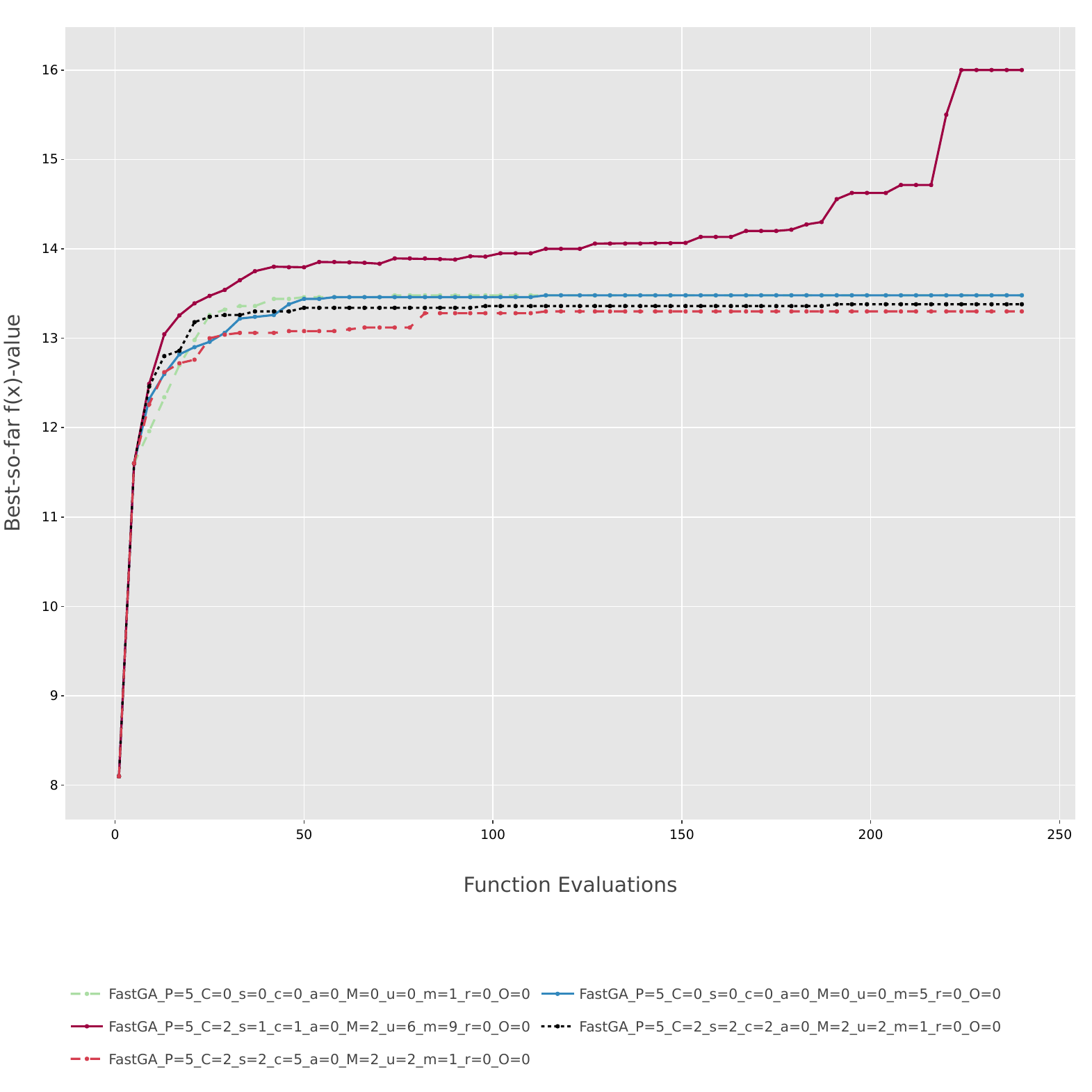}
    \caption{Convergence plot of baseline algorithms and elite, for problem 4.}
    \label{fig:convergence-4}
\end{figure}

\begin{figure}
    \centering
    \includegraphics[width=1.0\textwidth]{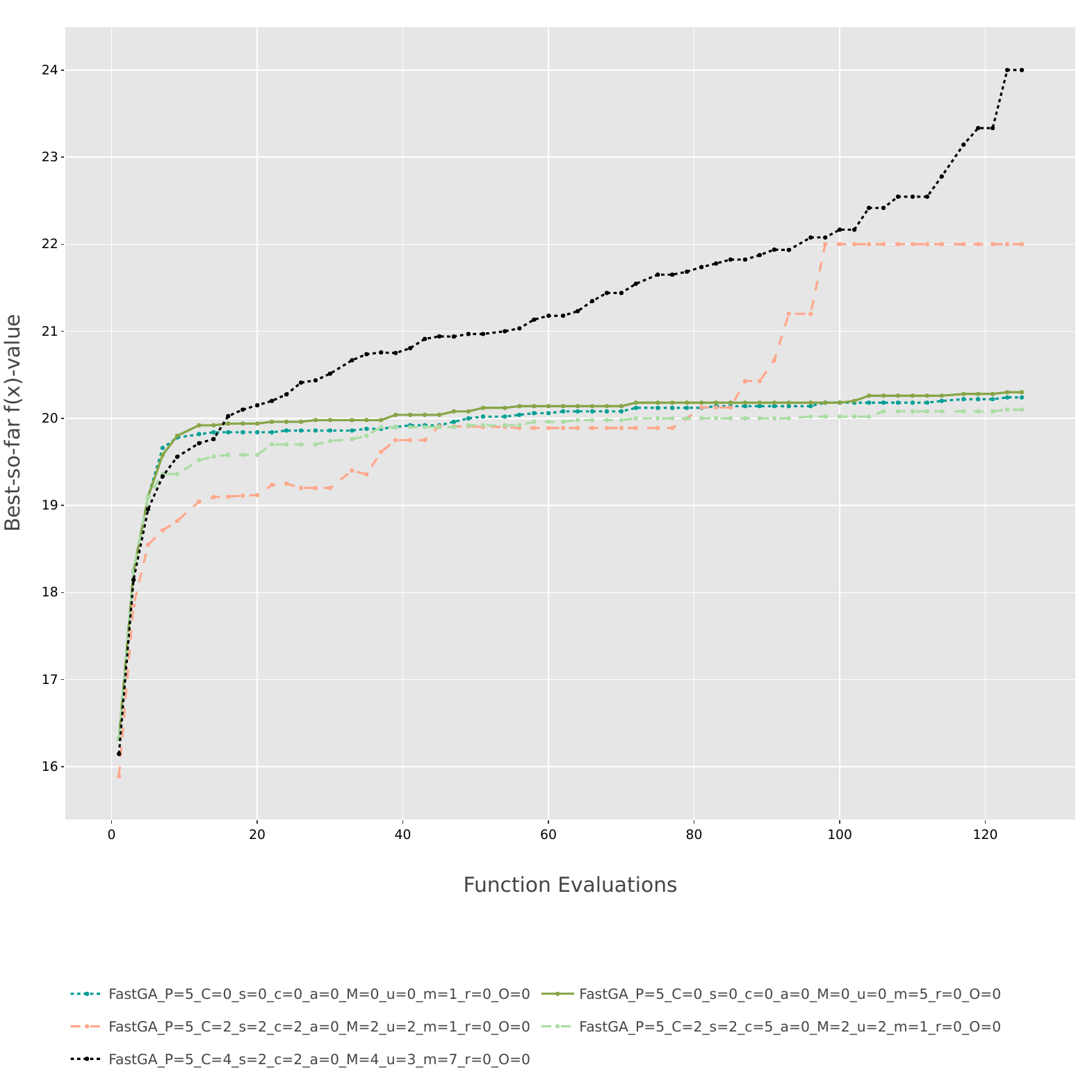}
    \caption{Convergence plot of baseline algorithms and elite, for problem 5.}
    \label{fig:convergence-5}
\end{figure}

\begin{figure}
    \centering
    \includegraphics[width=1.0\textwidth]{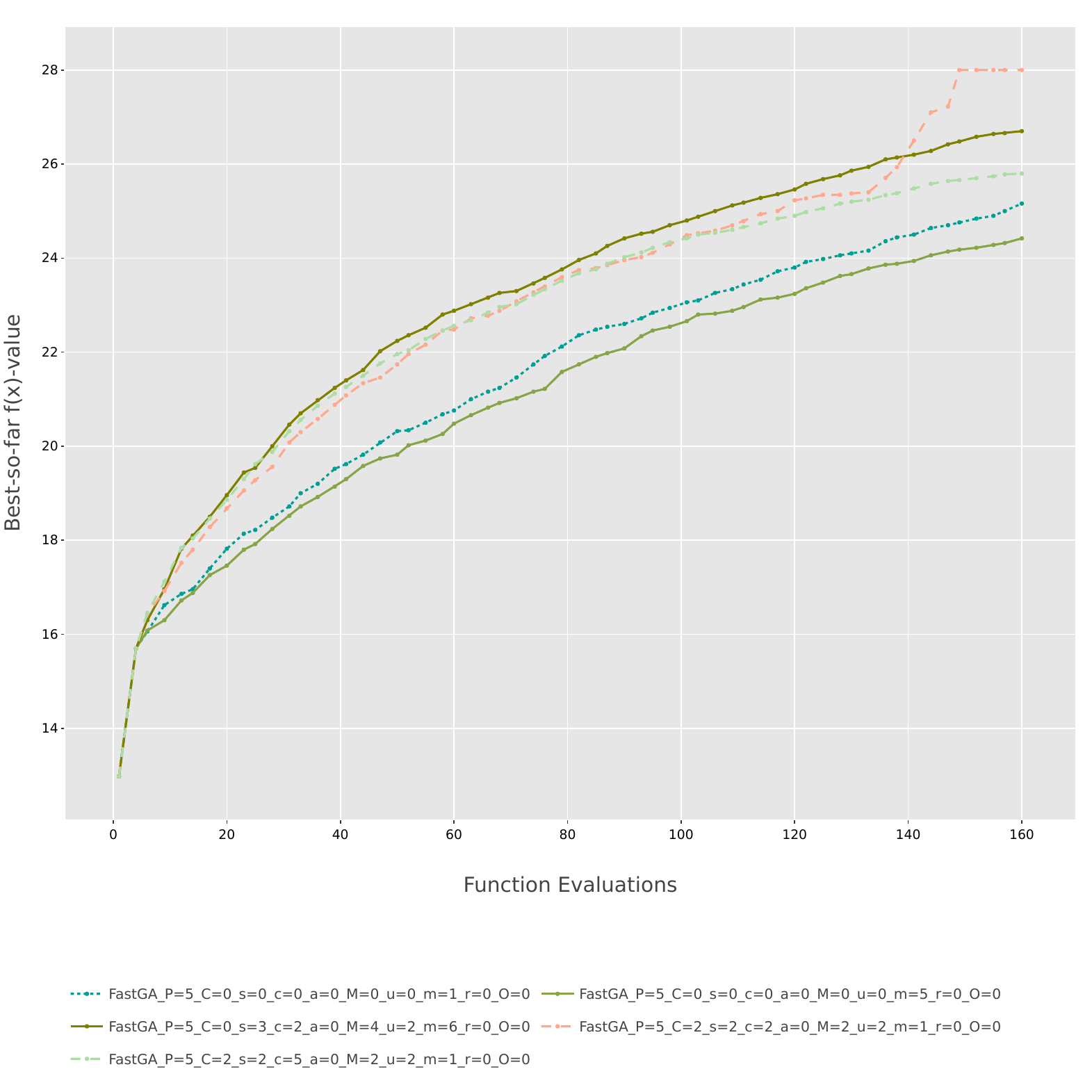}
    \caption{Convergence plot of baseline algorithms and elite, for problem 6.}
    \label{fig:convergence-6}
\end{figure}

\begin{figure}
    \centering
    \includegraphics[width=1.0\textwidth]{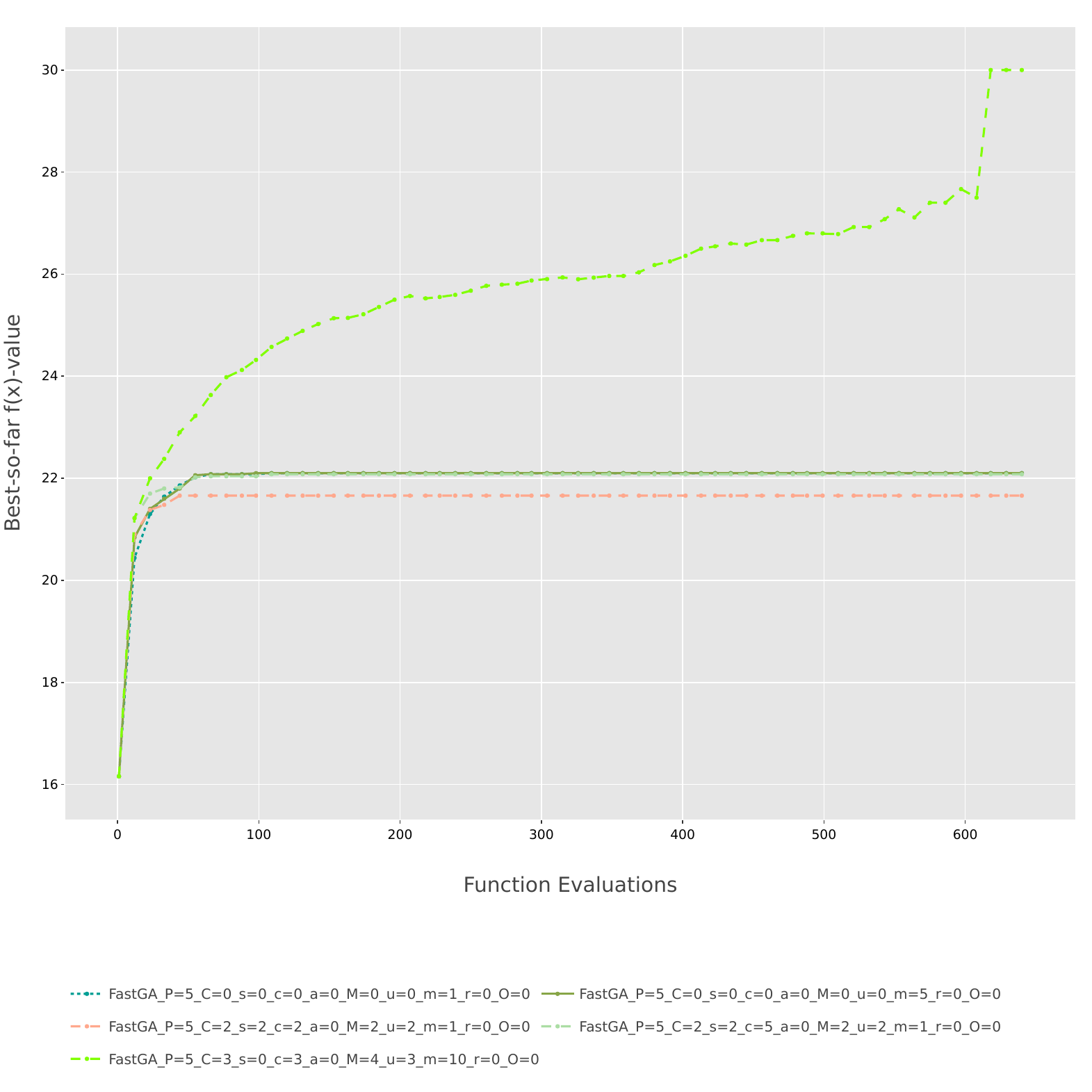}
    \caption{Convergence plot of baseline algorithms and elite, for problem 7.}
    \label{fig:convergence-7}
\end{figure}

\begin{figure}
    \centering
    \includegraphics[width=1.0\textwidth]{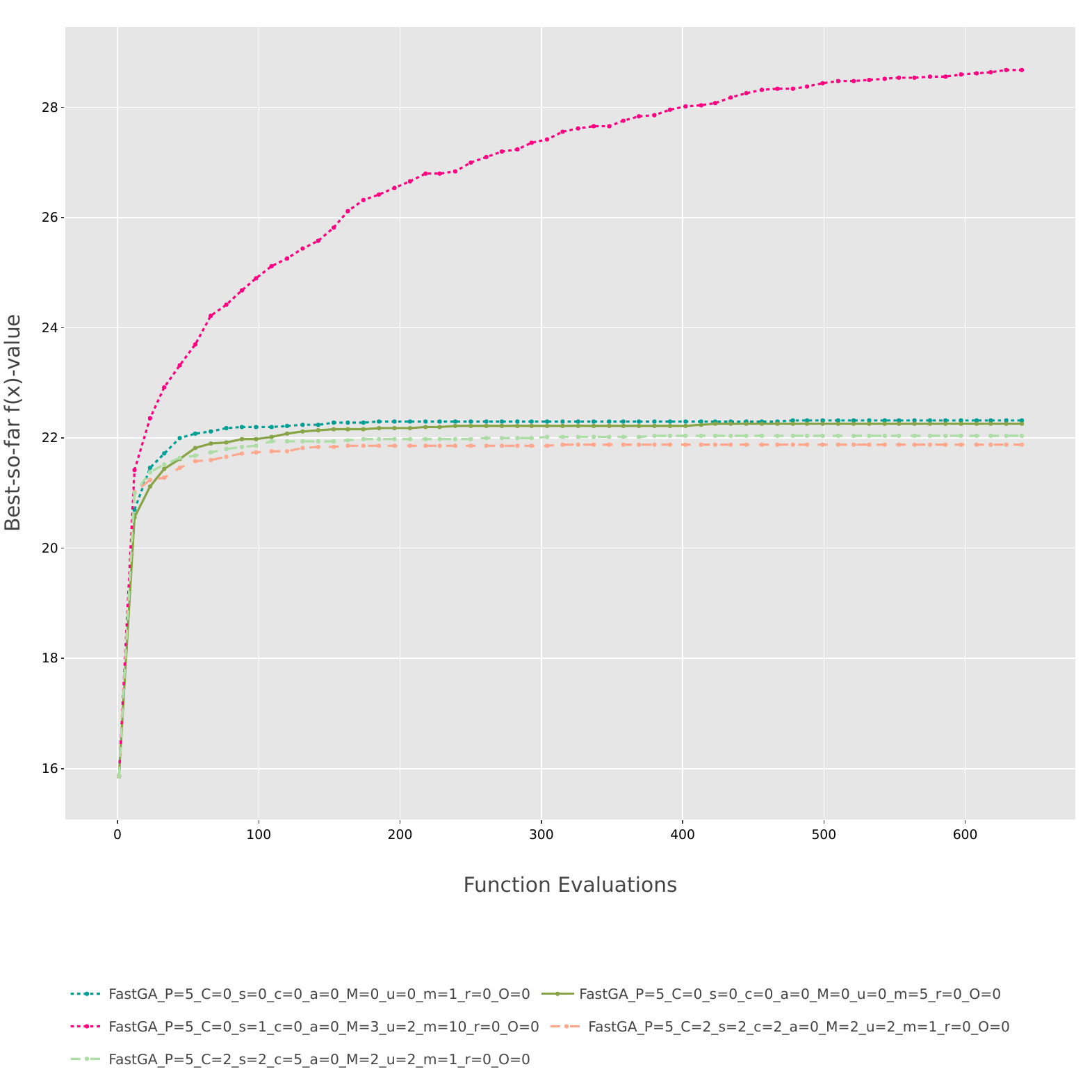}
    \caption{Convergence plot of baseline algorithms and elite, for problem 8.}
    \label{fig:convergence-8}
\end{figure}

\begin{figure}
    \centering
    \includegraphics[width=1.0\textwidth]{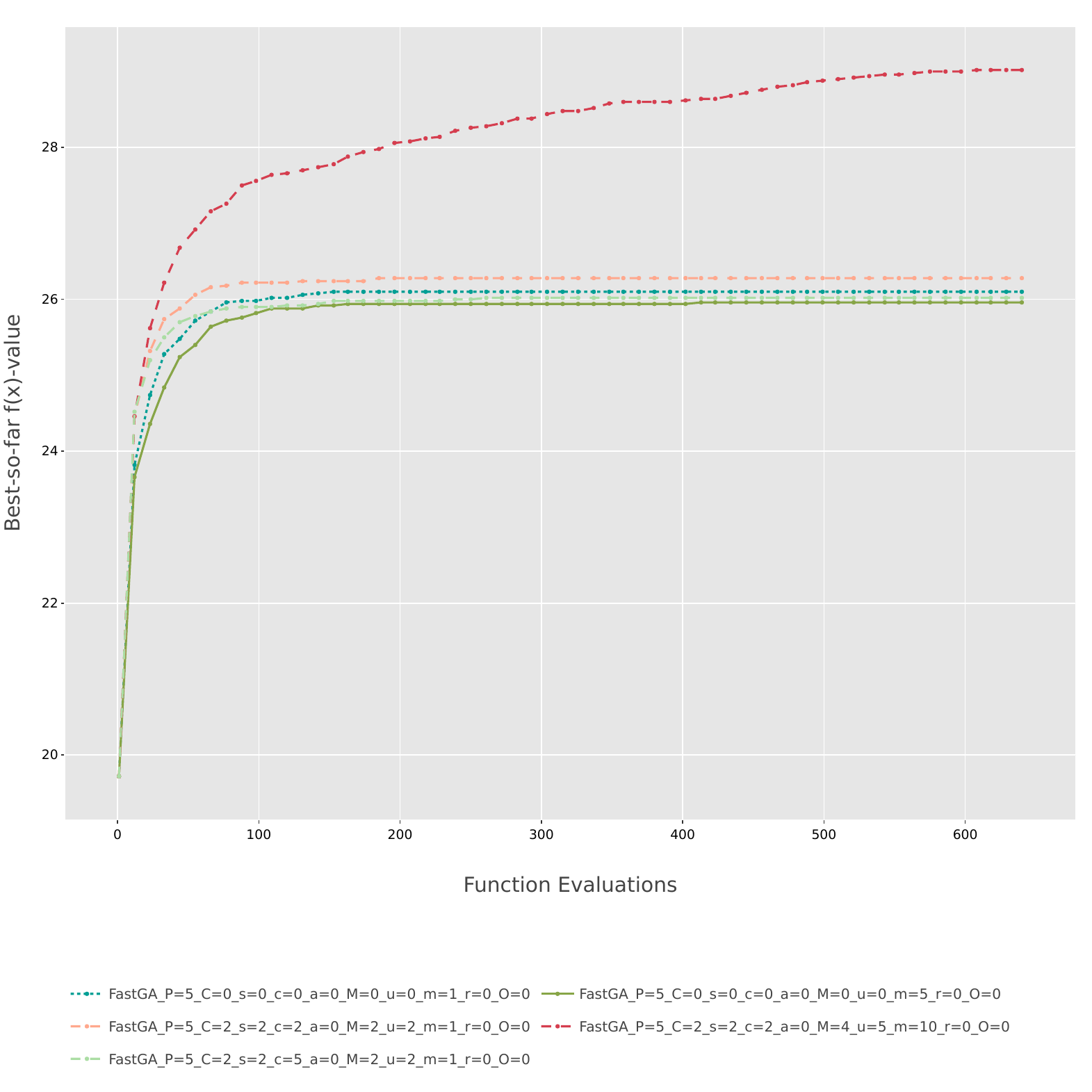}
    \caption{Convergence plot of baseline algorithms and elite, for problem 9.}
    \label{fig:convergence-9}
\end{figure}

\begin{figure}
    \centering
    \includegraphics[width=1.0\textwidth]{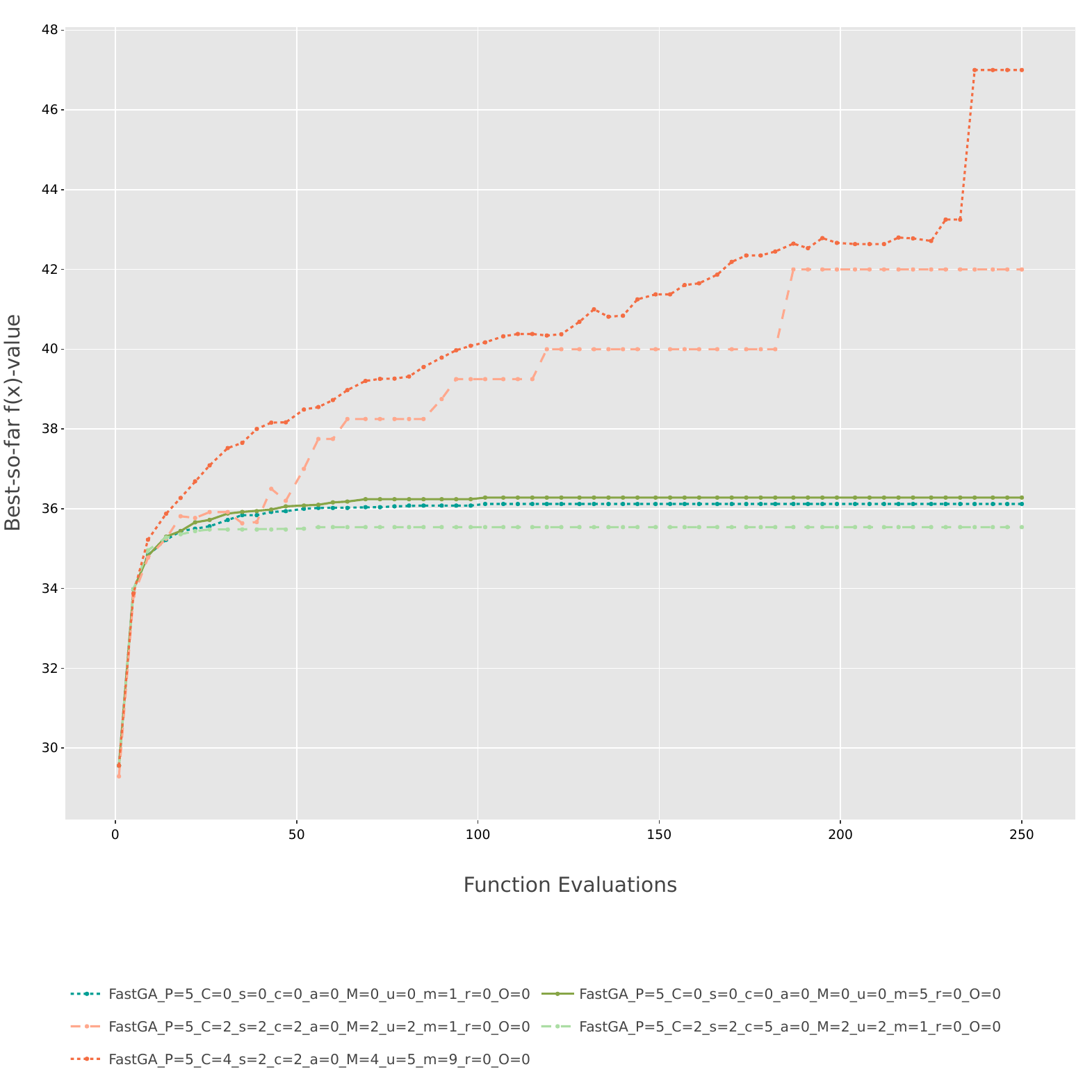}
    \caption{Convergence plot of baseline algorithms and elite, for problem 10.}
    \label{fig:convergence-10}
\end{figure}

\begin{figure}
    \centering
    \includegraphics[width=1.0\textwidth]{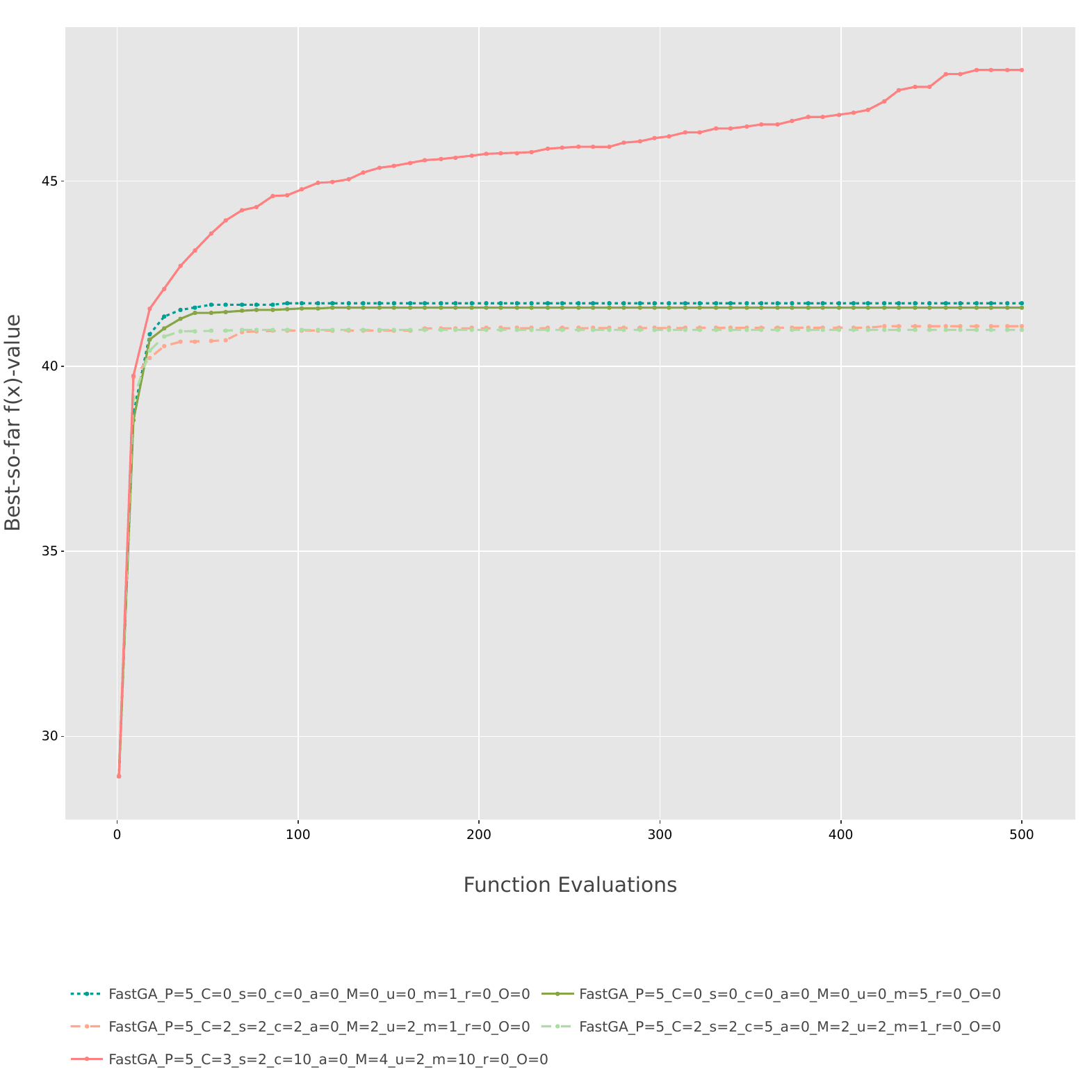}
    \caption{Convergence plot of baseline algorithms and elite, for problem 11.}
    \label{fig:convergence-11}
\end{figure}

\begin{figure}
    \centering
    \includegraphics[width=1.0\textwidth]{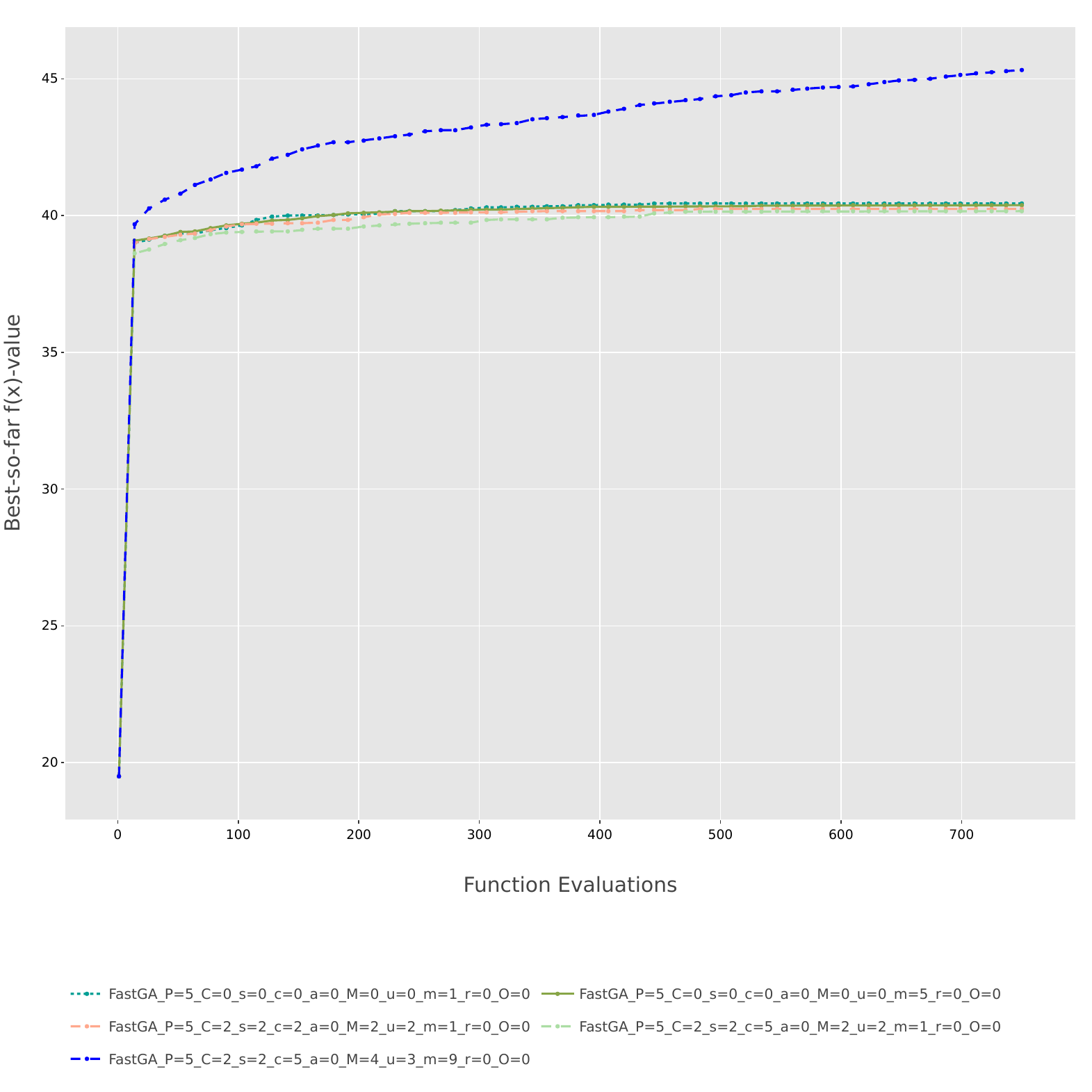}
    \caption{Convergence plot of baseline algorithms and elite, for problem 12.}
    \label{fig:convergence-12}
\end{figure}

\begin{figure}
    \centering
    \includegraphics[width=1.0\textwidth]{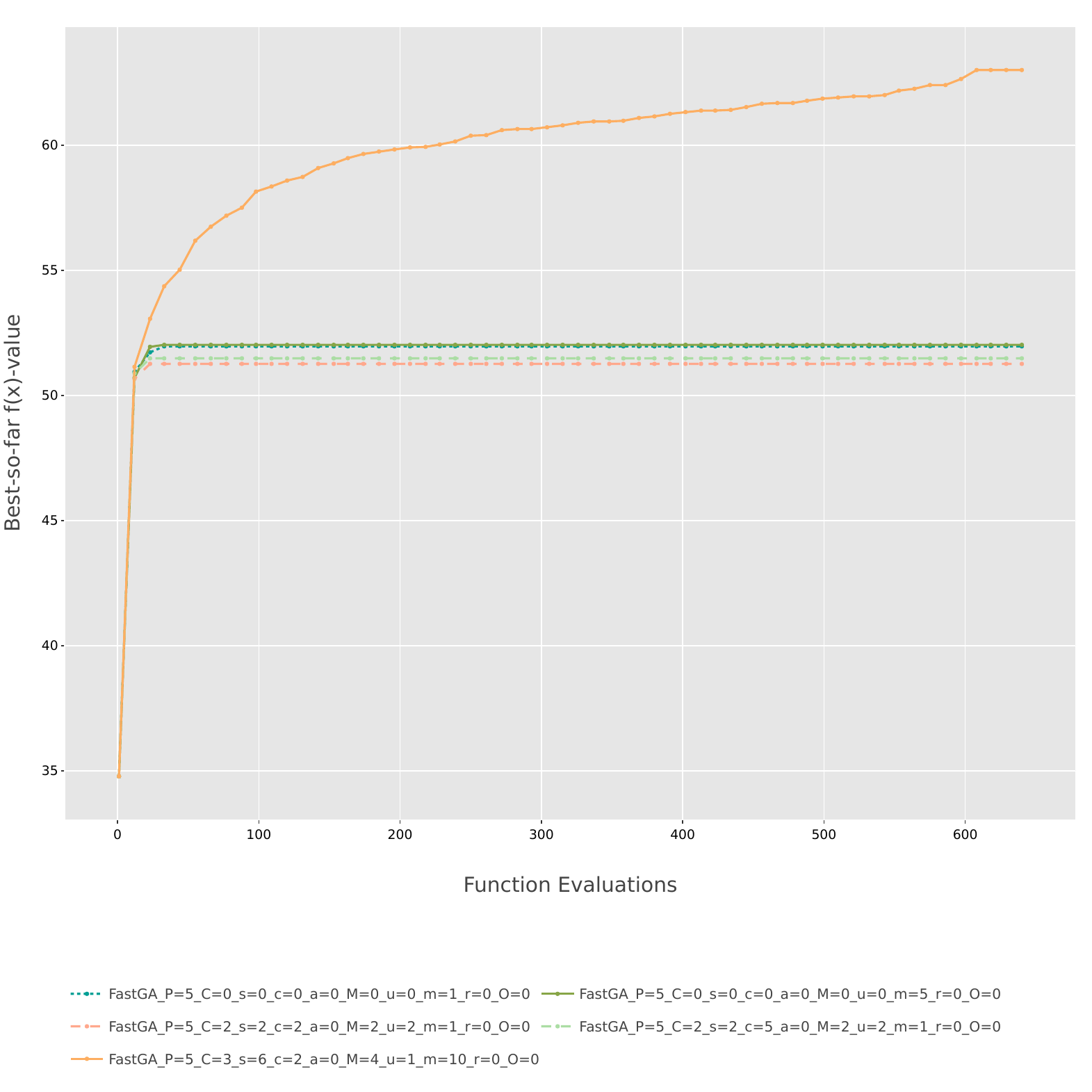}
    \caption{Convergence plot of baseline algorithms and elite, for problem 13.}
    \label{fig:convergence-13}
\end{figure}

\begin{figure}
    \centering
    \includegraphics[width=1.0\textwidth]{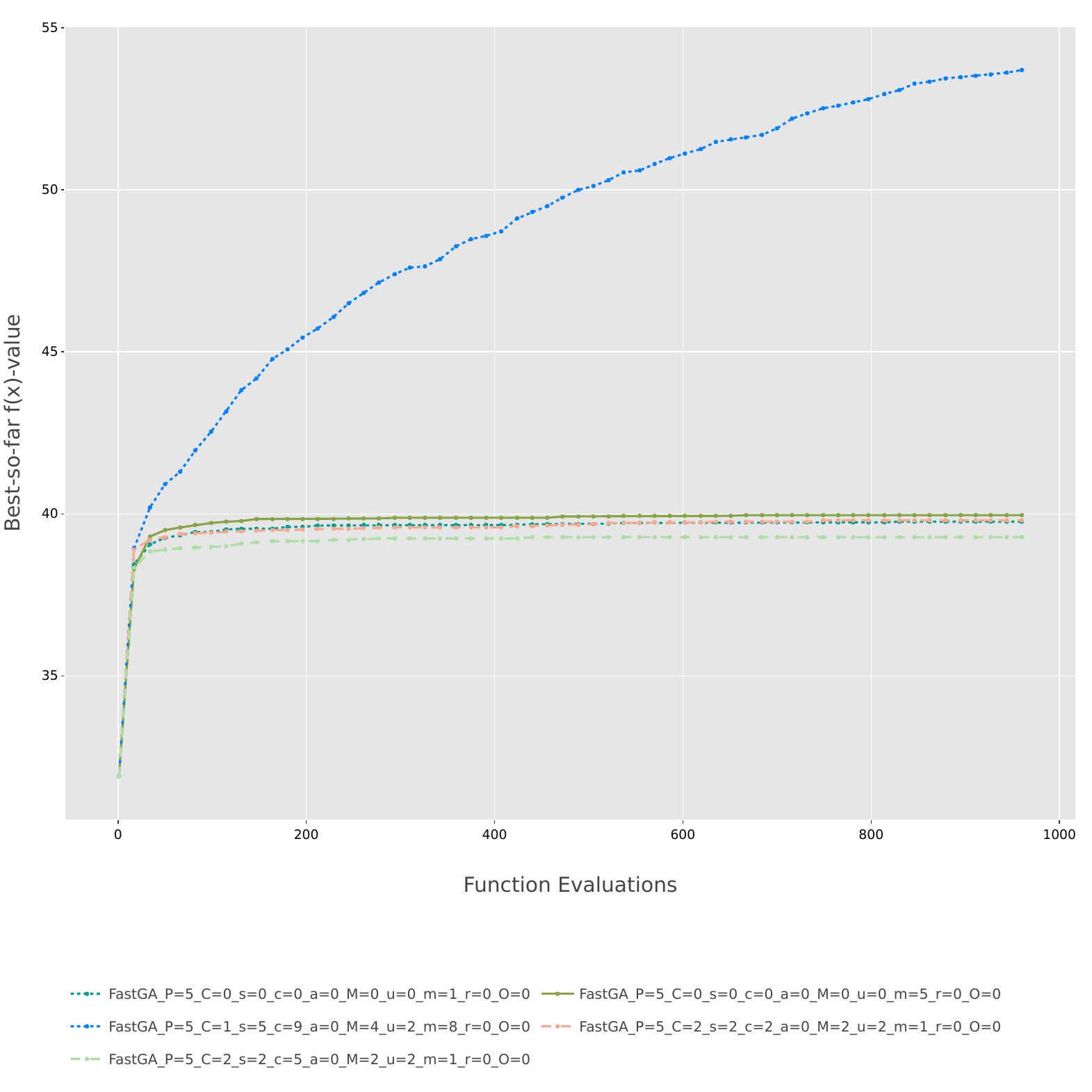}
    \caption{Convergence plot of baseline algorithms and elite, for problem 14.}
    \label{fig:convergence-14}
\end{figure}

\begin{figure}
    \centering
    \includegraphics[width=1.0\textwidth]{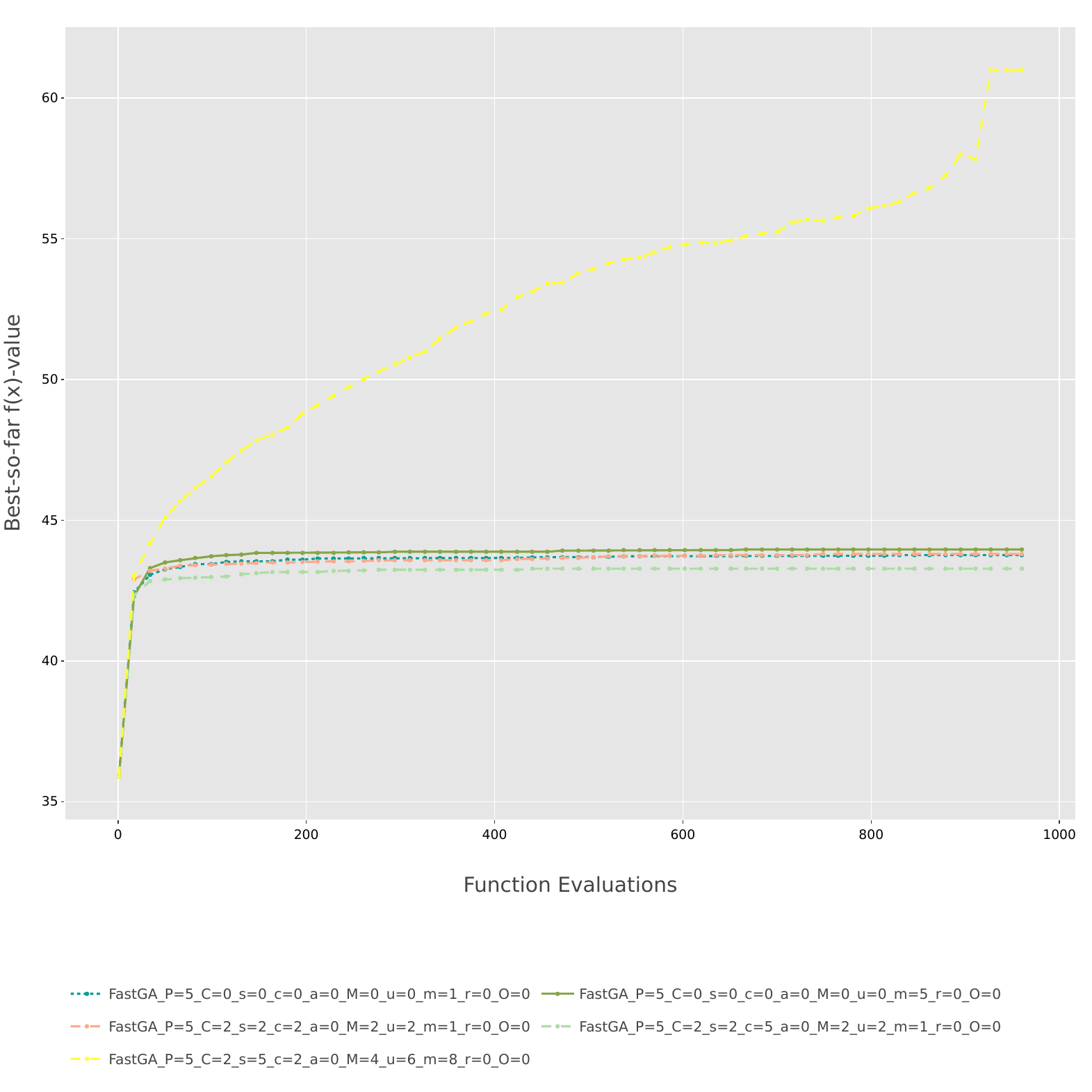}
    \caption{Convergence plot of baseline algorithms and elite, for problem 15.}
    \label{fig:convergence-15}
\end{figure}

\begin{figure}
    \centering
    \includegraphics[width=1.0\textwidth]{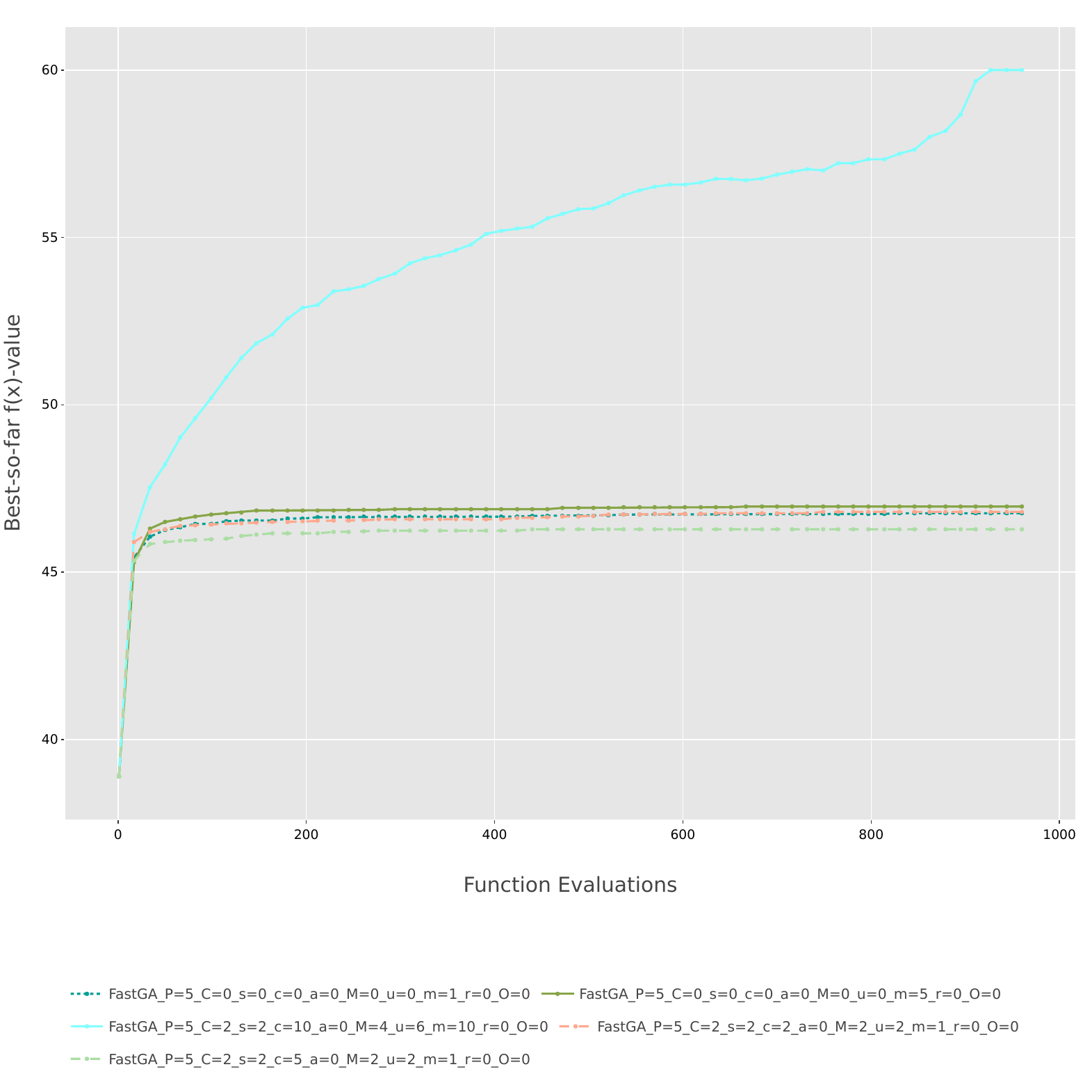}
    \caption{Convergence plot of baseline algorithms and elite, for problem 16.}
    \label{fig:convergence-16}
\end{figure}

\begin{figure}
    \centering
    \includegraphics[width=1.0\textwidth]{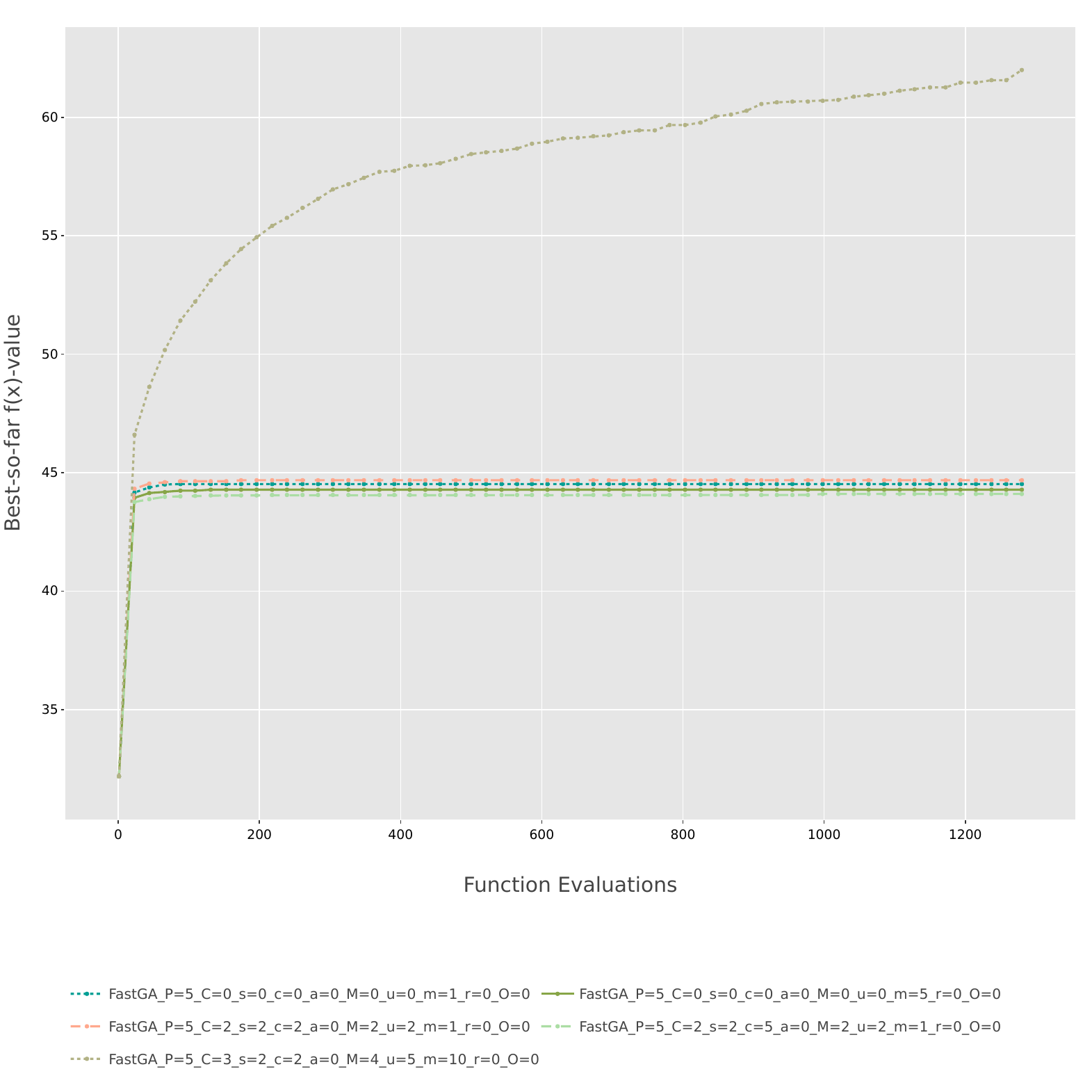}
    \caption{Convergence plot of baseline algorithms and elite, for problem 17.}
    \label{fig:convergence-17}
\end{figure}

\begin{figure}
    \centering
    \includegraphics[width=1.0\textwidth]{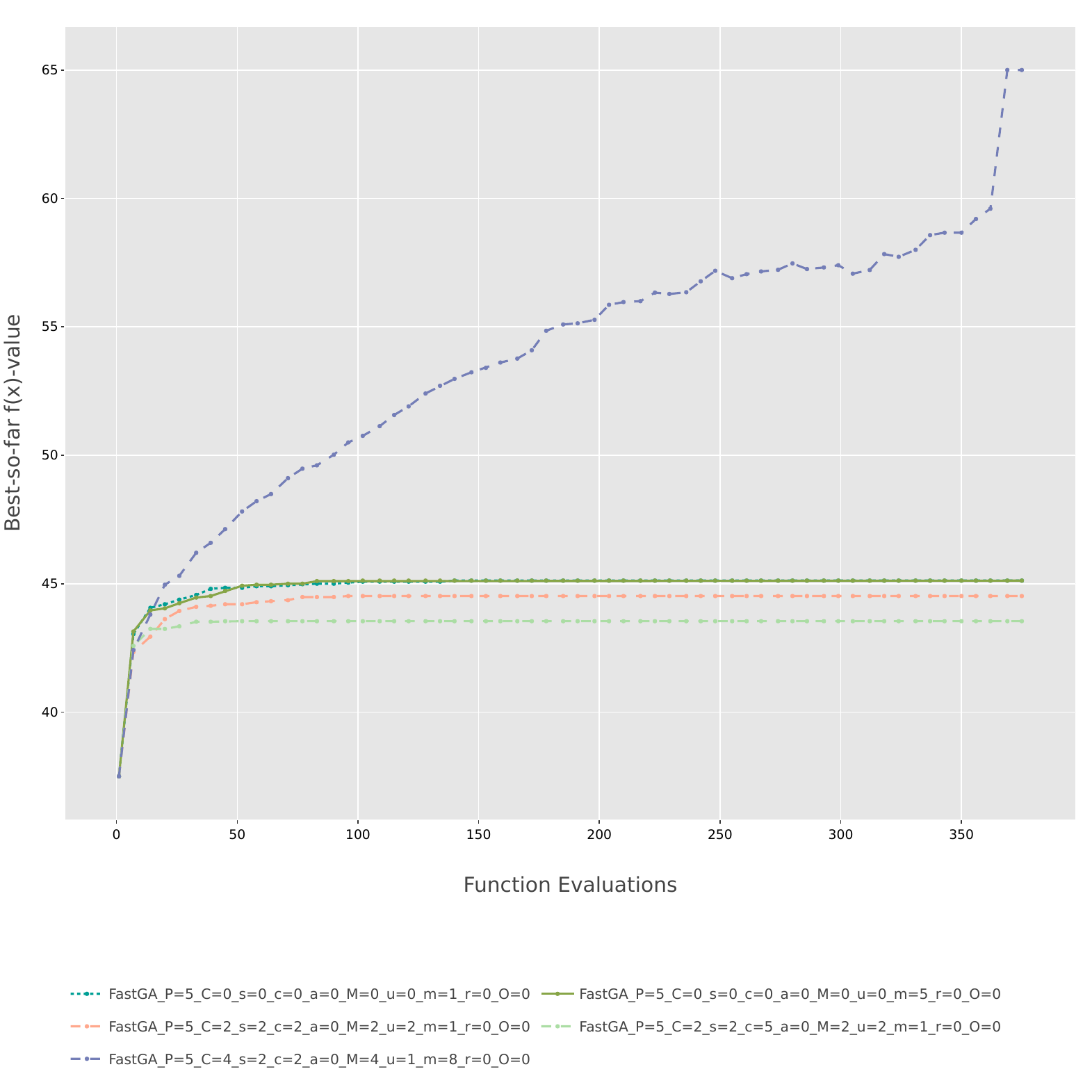}
    \caption{Convergence plot of baseline algorithms and elite, for problem 18.}
    \label{fig:convergence-18}
\end{figure}

\begin{figure}
    \centering
    \includegraphics[width=1.0\textwidth]{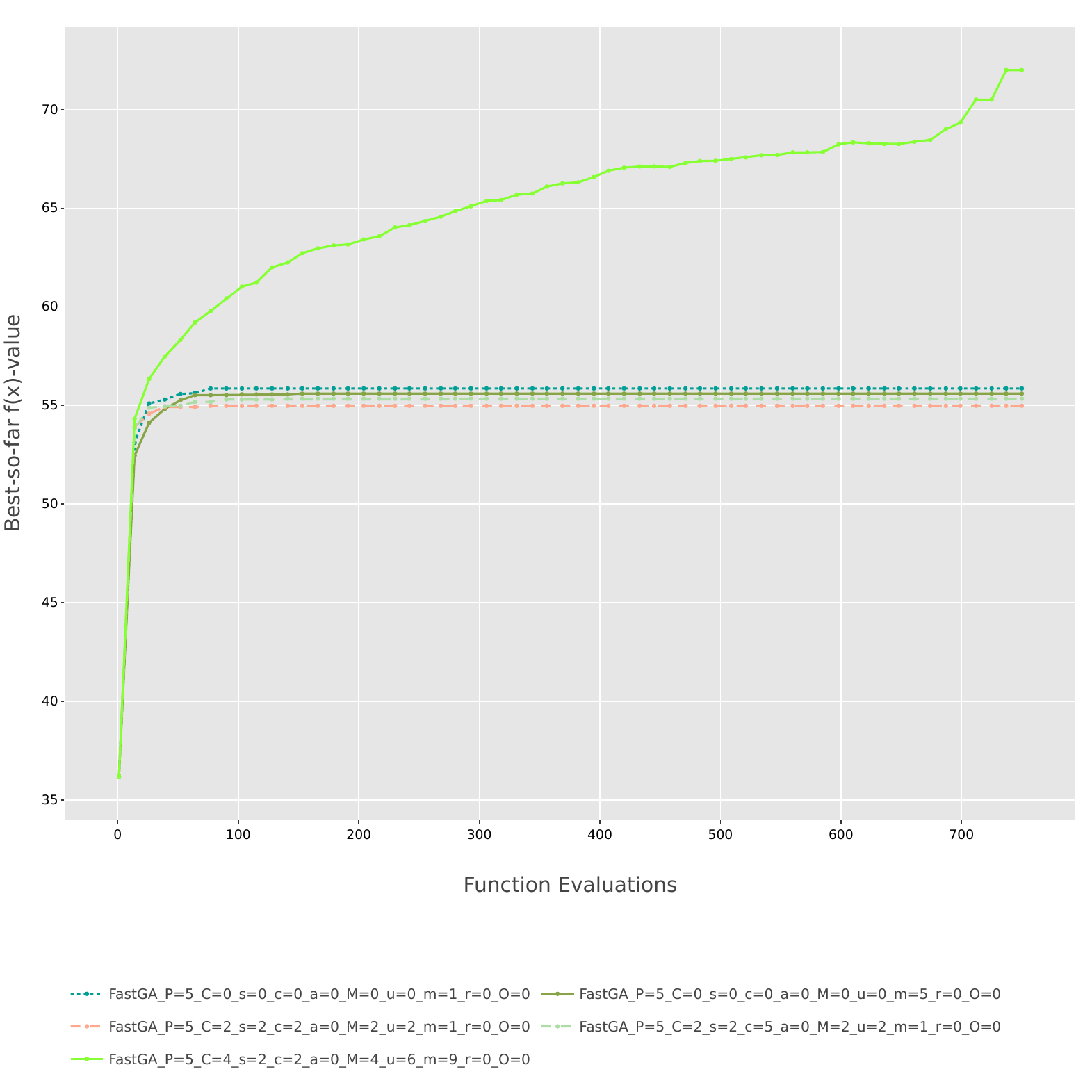}
    \caption{Convergence plot of baseline algorithms and elite, for problem 19.}
    \label{fig:convergence-19}
\end{figure}

\end{document}